\def\reviewclean{1}
\def\arxivversion{1}
\documentclass{article}
\usepackage[T1]{fontenc}
\usepackage{iclr2027_conference,times}
\usepackage{amssymb}
\usepackage{comment}

\newcommand{\TODO}[1]{\textbf{\color{blue}[yt: #1]}}

\usepackage{url}
\usepackage{wrapfig}
\usepackage{needspace}
\usepackage{capt-of}
\usepackage{graphicx}
\usepackage{multirow}
\usepackage{siunitx}
\usepackage{booktabs}
\usepackage{adjustbox}
\usepackage{tabularx}
\usepackage{colortbl}
\usepackage{xcolor}
\usepackage{bm}
\usepackage{float}
\usepackage{placeins}

\definecolor{best}{rgb}{0.96, 0.57, 0.58}
\definecolor{second}{rgb}{0.98, 0.78, 0.57}
\definecolor{third}{rgb}{1.0, 1.0, 0.56}

\usepackage{xcolor}
\usepackage{pifont}

\newcommand{\xmark}{\textcolor{red}{\ding{55}}}

\usepackage{amsmath}
\usepackage{algpseudocode}
\usepackage{setspace} 
\usepackage{makecell}

\newcommand{\syt}[1]{{\color{blue}[yt: #1]}}
\newcommand{\ltj}[1]{{\color{red}[ltj: #1]}}

\input{review-macros}
\usepackage{hyperref}
\hypersetup{hidelinks,pdftitle={Mira-Scene: Pixel-Aligned Layouts for Generative 3D Scene Reconstruction},pdfauthor={}}
\newcommand{\xjqi}[1]{\textcolor[rgb]{1.0,0,0}{{[\textbf{xjqi}: #1]}}}

\title{Mira-Scene: Pixel-Aligned Layouts\\
for Generative 3D Scene Reconstruction}
\author{Anonymous authors}
\ifdefined\arxivversion
\iclrfinalcopy
\reviewmarksfalse
\usepackage{marvosym}

\renewcommand{\xjqi}[1]{}
\renewcommand{\ltj}[1]{}
\renewcommand{\syt}[1]{}
\renewcommand{\TODO}[1]{}

\author{\hspace*{-\tabcolsep}\parbox{\textwidth}{\raggedright\normalfont\normalsize
\textbf{Yang-Tian Sun}\textsuperscript{1,*},
\textbf{Tianjia Liu}\textsuperscript{1,*},
\textbf{Zehuan Huang}\textsuperscript{2,\ensuremath{\dagger}},
\textbf{Yi-Hua Huang}\textsuperscript{1},
\textbf{Xiaoyang Lyu}\textsuperscript{1},\\[2pt]
\textbf{Ziyi Yang}\textsuperscript{1},
\textbf{Zi-Xin Zou}\textsuperscript{2},
\textbf{Yuan-Chen Guo}\textsuperscript{2},
\textbf{Yan-Pei Cao}\textsuperscript{2,\Letter},
\textbf{Xiaojuan Qi}\textsuperscript{1,\Letter}
}\hspace*{-\tabcolsep}}
\hypersetup{pdfauthor={Yang-Tian Sun, Tianjia Liu, Zehuan Huang, Yi-Hua Huang, Xiaoyang Lyu, Ziyi Yang, Zi-Xin Zou, Yuan-Chen Guo, Yan-Pei Cao, Xiaojuan Qi}}

\newcommand{\projectpageurl}{%
  \href{https://sunyangtian.github.io/Mira-Scene-web/}%
  {\underline{\textit{https://sunyangtian.github.io/Mira-Scene-web/}}}%
}

\makeatletter
\newcommand{\arxivauthorfootnote}{%
  \begingroup
  \renewcommand{\thefootnote}{}%
  \long\def\@makefntext##1{\noindent##1}%
  \footnotetext{\textsuperscript{1}The University of Hong Kong; \textsuperscript{2}VAST.\quad
  Project page: \projectpageurl.\\
  * Equal Contribution; \ensuremath{\dagger} Project Lead; \Letter\ Corresponding Authors.}%
  \endgroup
}
\makeatother
\fi

\begin{document}
\maketitle
\ifdefined\arxivversion\lhead{Preprint}\arxivauthorfootnote\vspace{-14pt}\fi

\begin{figure}[htbp]
\centering
\ifdefined\arxivversion
\includegraphics[width=\linewidth]{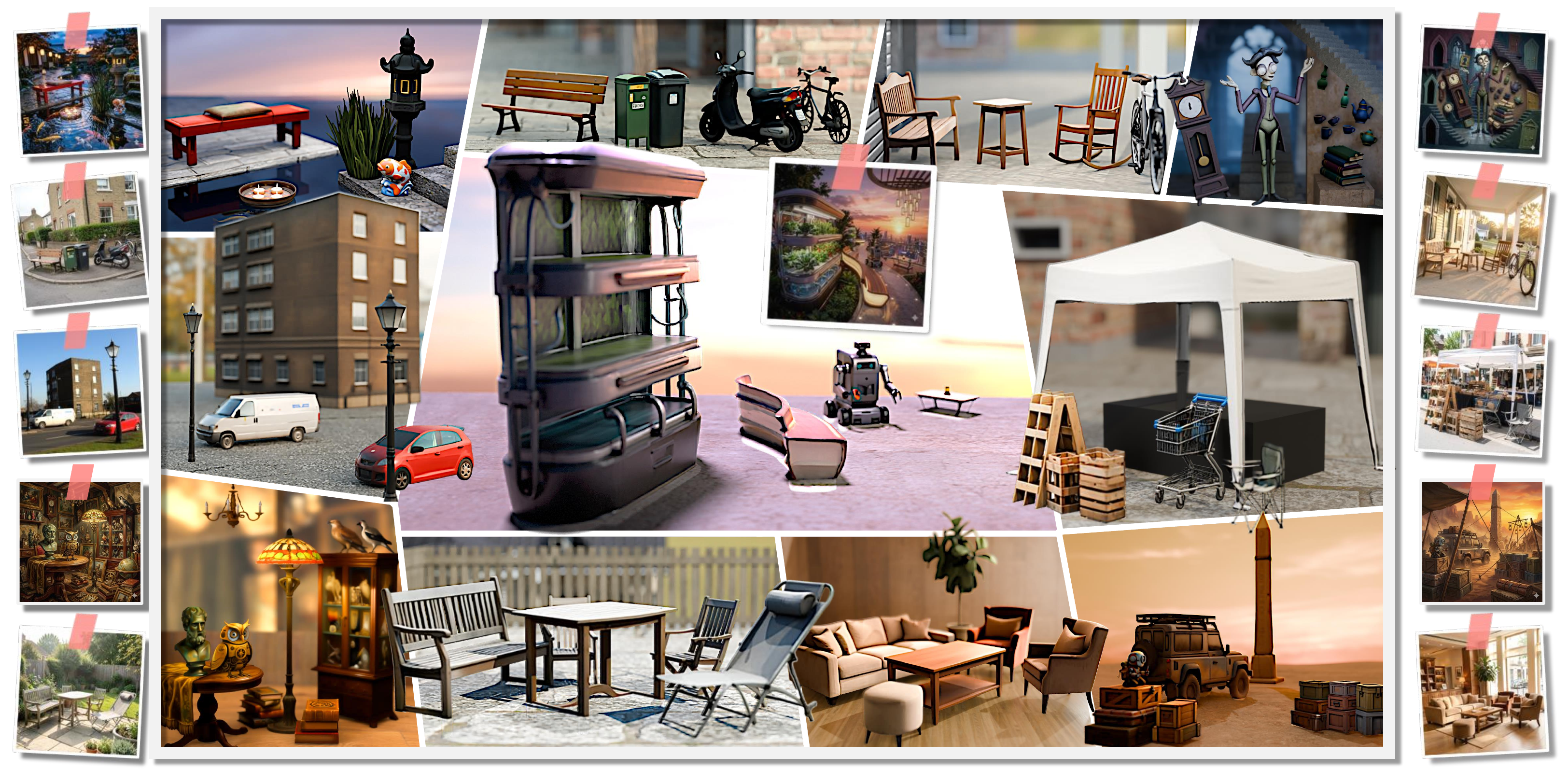}
\else
\includegraphics[width=\linewidth,height=0.62\textheight,keepaspectratio]{images/teaser.pdf}
\fi
\caption{Mira-Scene reconstructs compositional 3D scenes from a single image through canonical object generation and dense CCM--PCM correspondence alignment, preserving object-level detail and coherent layouts across indoor, outdoor, synthetic, and in-the-wild inputs.}
\label{fig:teaser}
\end{figure}

\begin{abstract}
Single-image 3D object generation can now produce high-fidelity assets, yet accurately placing them into a coherent scene layout remains an open challenge. A central difficulty lies in how object layout is represented. Holistic methods absorb placement into a scene-level generation process, sacrificing object-level detail. Compositional methods preserve object fidelity by decoupling geometry from layout, but typically parameterize layout as sparse, unbounded pose variables that are difficult to learn and generalize poorly under scarce scene-level supervision.
We present \textbf{Mira-Scene}, a compositional 3D scene reconstruction framework that replaces sparse pose regression with dense, bounded correspondence recovery. At its core is the \textbf{Canonical Coordinate Map (CCM)}, a pixel-aligned field that maps each visible object pixel to a surface coordinate in the object's bounded canonical space. When paired with a scene-space \textbf{Point Cloud Map (PCM)} from monocular geometry estimation, CCM induces dense canonical-to-scene correspondences from which object transformations are recovered through robust geometric alignment. Because CCM operates in bounded canonical space, it provides a stable prediction target that can be trained from scalable object-level 3D data without requiring scene-level layout annotations. Mira-Scene further introduces a multimodal diffusion transformer that jointly generates object geometry and CCMs, using modality-specific expert streams with shared attention and positional encoding to promote geometry-layout consistency. Experiments on indoor, outdoor, synthetic, and in-the-wild scenes show that Mira-Scene substantially outperforms strong baselines in layout accuracy, \reviewadded{achieving relative gains of} \reviewadded{39.8\%} in 3D-IoU and \reviewadded{16.5\%} in 2D-IoU over SAM3D, \reviewadded{using limited open-source training data.}

\end{abstract}

\section{Introduction}
\label{sec:intro}
\reviewadded{Reconstructing compositional 3D scenes from a single image requires recovering both high-fidelity object geometry and accurate object placement in a shared scene coordinate frame. This capability supports editable 3D content creation, embodied simulation, AR/VR, and robotic interaction. Recent advances in single-image 3D object generation~\citep{hong2023lrm, xiang2025structured, li2025triposg, zhang2024clay} produce high-fidelity assets, but these models operate in a canonical object space and do not reason about where each object should be placed in the scene.}

\begin{wrapfigure}{r}{0.50\textwidth}
\centering
  \includegraphics[width=\linewidth]{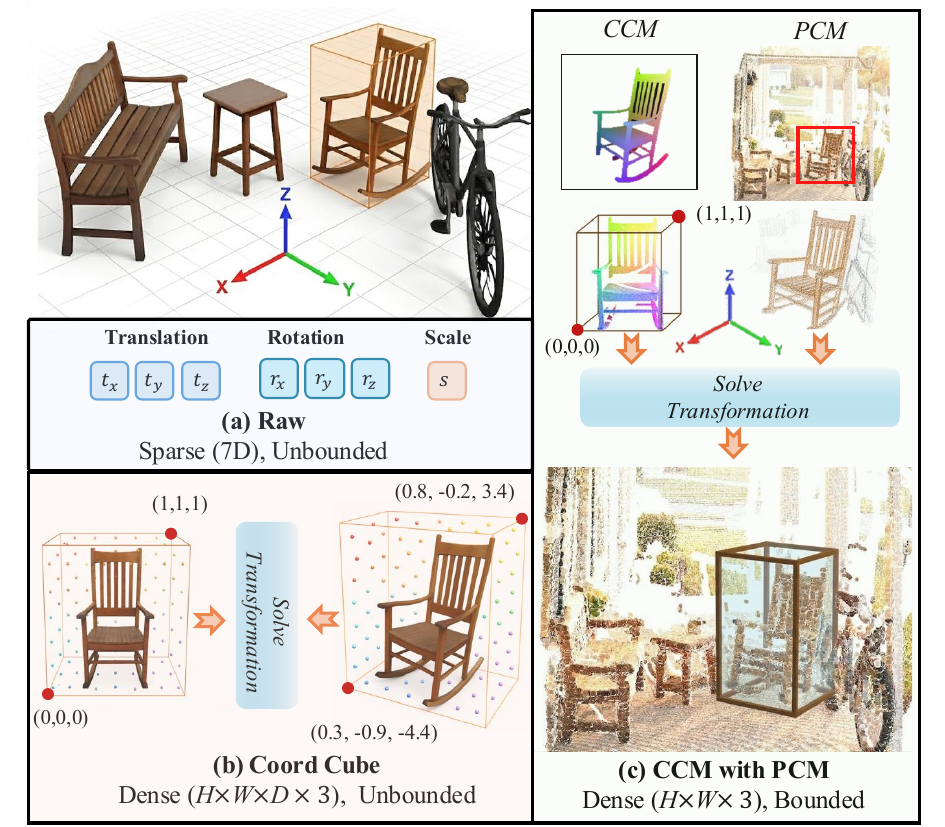}
  \caption{Comparison of layout representations. (a) Raw pose regression is sparse and unbounded; (b) Coord Cube densifies prediction but still regresses scene-space coordinates; (c) Our CCM predicts dense canonical-space correspondences and recovers object placement by aligning them with a scene-space PCM.}
  \label{fig:layout_representation}
\end{wrapfigure}

A central question for single-image compositional 3D scene generation is how to represent the object layout, especially given the scarcity of 3D scene data compared with abundant object-level 3D assets~\citep{deitke2023objaverse}. Treating the entire scene as a single holistic 3D asset~\citep{huang2025midi, ling2025scene, lin2025partcrafter, wang2026scenetransporter} can be viewed as an implicit layout representation: object placement is absorbed into a unified scene-level generation process, benefiting from strong priors learned by object-level generative models. However, under a fixed token, voxel, or latent budget, the representation must cover the full spatial extent of the scene, leaving fewer effective degrees of freedom for each object. Fine structures, small objects, and object boundaries are often under-resolved, making such representations less suitable for object-level editing, simulation, and interaction. We therefore focus on compositional representations that explicitly decouple high-resolution object geometry in canonical space from object layout in scene space (Fig.~\ref{fig:layout_representation}). This decomposition preserves object fidelity, but the central difficulty shifts to layout recovery. 

\AddToHookNext{cmd/@makecol/before}{\setlength{\textfloatsep}{12pt}}

The most common layout representation directly parameterizes each object's placement as translation, rotation, and scale. As illustrated in Fig.~\ref{fig:layout_representation}(a), these parameters form a sparse and unbounded target that is difficult for neural networks to regress accurately, especially under occlusion, perspective ambiguity, and long-tailed configurations. \reviewadded{Scarce scene-level supervision compounds this difficulty.} Even data-centric systems such as SAM3D~\citep{chen2026sam}, which construct large-scale data engines with professional artist intervention, still express layout through sparse pose variables and \reviewadded{have limited object alignment accuracy}. This suggests that data scaling alone cannot fully address the representation difficulty for general and accurate layout prediction.

A natural way to reduce sparsity is to densify scene-space prediction. We consider a \textbf{Coord Cube} representation, shown in Fig.~\ref{fig:layout_representation}(b), where the model predicts the scene-space locations of uniformly sampled points in the object's canonical space. The object transformation is then recovered by aligning these predicted scene-space points with their canonical coordinates. As confirmed by our ablation study (Tab.~\ref{tab:layout_comparison}), Coord Cube improves over raw pose regression by providing a denser target that is more robust to local prediction errors. However, the predicted coordinates remain unbounded scene-space quantities.
Under limited scene data, the robustness gains of Coord Cube remain constrained.

\reviewadded{These observations motivate a layout representation that is both dense and bounded. We therefore propose \textbf{Mira-Scene}, a generative compositional 3D reconstruction framework that replaces sparse pose or unbounded point prediction with dense and bounded correspondence recovery. At its core is the \textbf{Canonical Coordinate Map} (CCM), a pixel-aligned field that maps each visible object pixel to a surface coordinate in the object's bounded canonical space, as shown in Fig.~\ref{fig:layout_representation}(c). When paired with a scene-space \textbf{Point Cloud Map} (PCM) estimated by a monocular geometry model~\citep{wang2025moge, xu2025pixel}, the CCM induces dense canonical-to-scene correspondences, allowing object transformations to be recovered through robust geometric alignment rather than direct neural regression. The bounded canonical coordinates provide a stable target for scalable object-level supervision, \reviewadded{while dense correspondences improve robustness to local prediction errors.}}

Further, since object geometry and CCMs are both defined in the same canonical object space, they can be generated coherently within a unified framework to enforce consistency between shape and layout. We therefore introduce a multimodal diffusion transformer for geometry-layout co-generation. The model represents canonical 3D geometry and 2D CCMs as two modality-specific streams, while enabling information exchange through shared self-attention. This design preserves the distinct structures of 3D geometry and pixel-aligned coordinate maps, while promoting consistency between the generated shape and its corresponding layout representation. We further introduce a shared positional encoding strategy that embeds geometry and layout tokens into a common positional space, enabling more effective cross-modal interaction during generation. 

\reviewadded{Experiments on indoor, outdoor, synthetic, and in-the-wild scenes show that \textbf{Mira-Scene} produces detailed object geometry and substantially more accurate scene layouts than strong baselines (see Fig.~\ref{fig:teaser}). On \cite{blendswap} benchmark, Mira-Scene improves 3D-IoU from \textbf{0.520} to \textbf{0.727} and 2D-IoU from \textbf{0.672} to \textbf{0.783} over SAM3D, \reviewadded{using substantially less, publicly sourced training data.}}

\suppressfloats[t]
In summary, our contributions are:
\begin{itemize}
    \setlength{\itemsep}{2pt}
    \setlength{\parsep}{0pt}
    \setlength{\topsep}{4pt}
    \item We introduce CCM with PCM, a pixel-aligned correspondence-based layout representation that replaces sparse object pose regression with robust geometric alignment.
    \item We present a geometry-layout co-generation model that jointly predicts canonical object geometry and pixel-aligned CCMs using a multimodal diffusion transformer.
    \item We demonstrate data-efficient compositional scene reconstruction across indoor, outdoor, and in-the-wild scenes, achieving substantially better layout accuracy than strong baselines trained with larger-scale supervision.
\end{itemize}


\section{Method}

\begin{figure}[t]
\centering
\includegraphics[width=\textwidth,height=0.72\textheight,keepaspectratio]{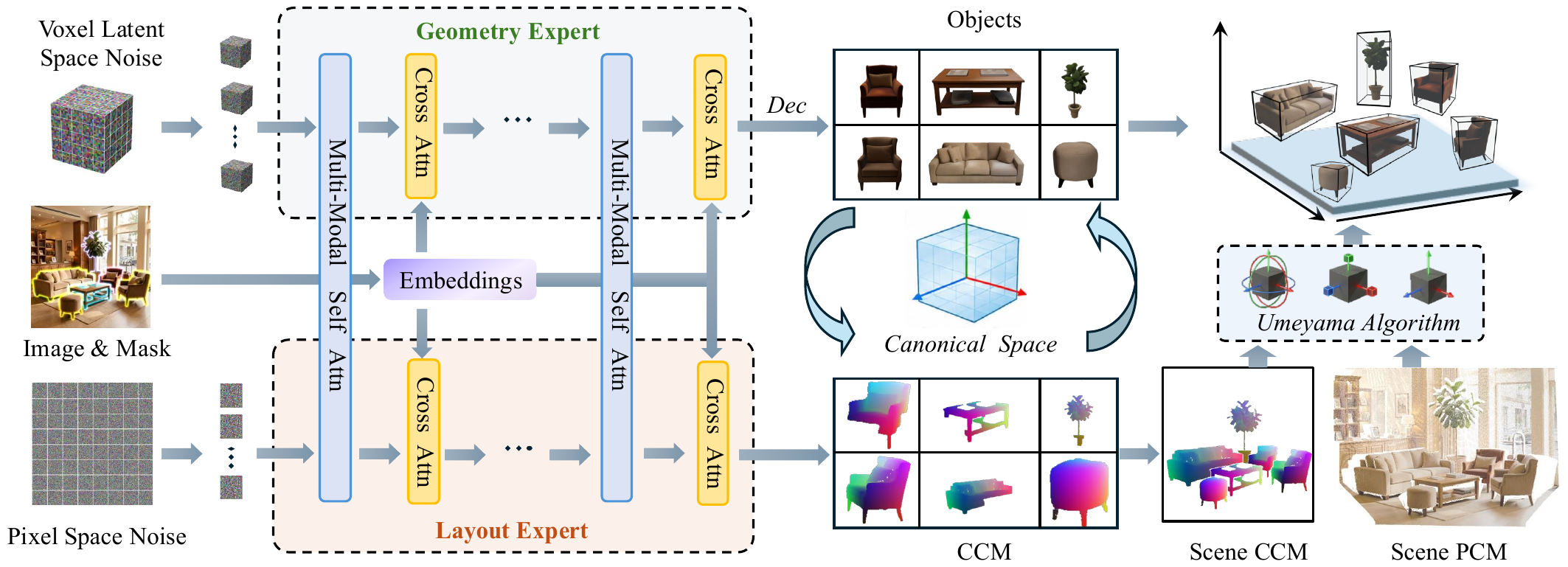}
  \caption{Overview of Mira-Scene. Given a single image and object masks, Mira-Scene jointly generates canonical object geometry and pixel-aligned CCMs for each object. Object placement is recovered by aligning the predicted CCMs with a scene-space PCM (Point Cloud Map) estimated from monocular geometry, \reviewadded{yielding object-level assets in a coherent scene layout.}}
  \label{fig:pipeline}
\end{figure} 

\subsection{Problem Formulation}
\label{sec:problem_formulation}

Given an input image $I \in \mathbb{R}^{H\times W\times 3}$ and a set of instance masks $\{M_k\}_{k=1}^{K}$, our goal is to reconstruct a compositional 3D scene represented as a set of posed object assets $\{(S_k,T_k)\}_{k=1}^{K}$. Each object geometry $S_k$ is generated in a canonical object space, and $T_k=(s_k,R_k,t_k)$ maps it to the scene coordinate frame, where $s_k\in\mathbb{R}_{+}$ is an isotropic scale, $R_k\in SO(3)$ is a rotation, and $t_k\in\mathbb{R}^{3}$ is a translation. We use the camera coordinate frame of the input image as the scene frame. {When masks are not supplied, they can be obtained using a VLM-guided agentic segmentation pipeline built on SAM3~\citep{carion2026sam}, as detailed in Appendix~\ref{app:auto_segmentation}.}

Instead of directly regressing $T_k$, \textbf{Mira-Scene} predicts the canonical geometry $S_k$ and a \emph{Canonical Coordinate Map (CCM)} $C_k$ for each object. The explicit transformation $T_k$ is recovered afterwards by aligning $C_k$ with a scene-space \emph{Point Cloud Map (PCM)}, as described in Sec.~\ref{sec:scene_assemble}. 
Therefore, our per-object generative objective is \reviewadded{$q(S_k,C_k \mid I,M_k)$},
\reviewadded{jointly predicting object-level geometry and its dense layout representation from image and mask.}

\subsection{Pixel-Aligned Layout Representation}
\label{sec:dense_corr}

The central representation in Mira-Scene is CCM with PCM. We normalize each object into a bounded canonical coordinate system. For object $k$, the CCM $C_k$ is predicted in the cropped object image space. For each visible pixel $u$ inside the object mask, $C_k(u)\in\mathbb{R}^{3}$ stores the canonical coordinate of the object surface point observed at that pixel. Background pixels and invalid pixels are excluded by a validity mask. Although $C_k$ is stored as a three-channel image, its channels represent canonical $xyz$ coordinates rather than color.

The PCM $P\in\mathbb{R}^{H\times W\times 3}$ is a dense point map in the scene frame, where $P(u)$ gives the 3D scene point observed at pixel $u$. In practice, $P$ can be obtained from a depth camera or estimated by monocular geometry prediction~\citep{wang2025moge, xu2025pixel}. After pasting the crop-space CCM back to the full image according to the object crop and mask, 
each valid pixel provides a dense correspondence \reviewadded{$C_k(u) \leftrightarrow P(u)$} between the object's canonical space and the scene frame. The object-to-scene transformation can then be recovered by geometric alignment rather than neural pose regression.

This representation differs from scene-space coordinate prediction such as Coord Cube in Fig.~\ref{fig:layout_representation}. CCM coordinates live in a bounded canonical object space and can be supervised from rendered object assets without requiring ground-truth scene-level layouts. \reviewadded{CCM thus enables scalable object-level pre-training and accurate scene-space placement through PCM alignment.}

\subsection{Geometry-Layout Co-Generation}
\label{sec:co_gen}
\begin{wrapfigure}{R}{0.50\textwidth}
\setlength{\abovecaptionskip}{5pt}
\color{black}
\centering
  \includegraphics[width=\linewidth]{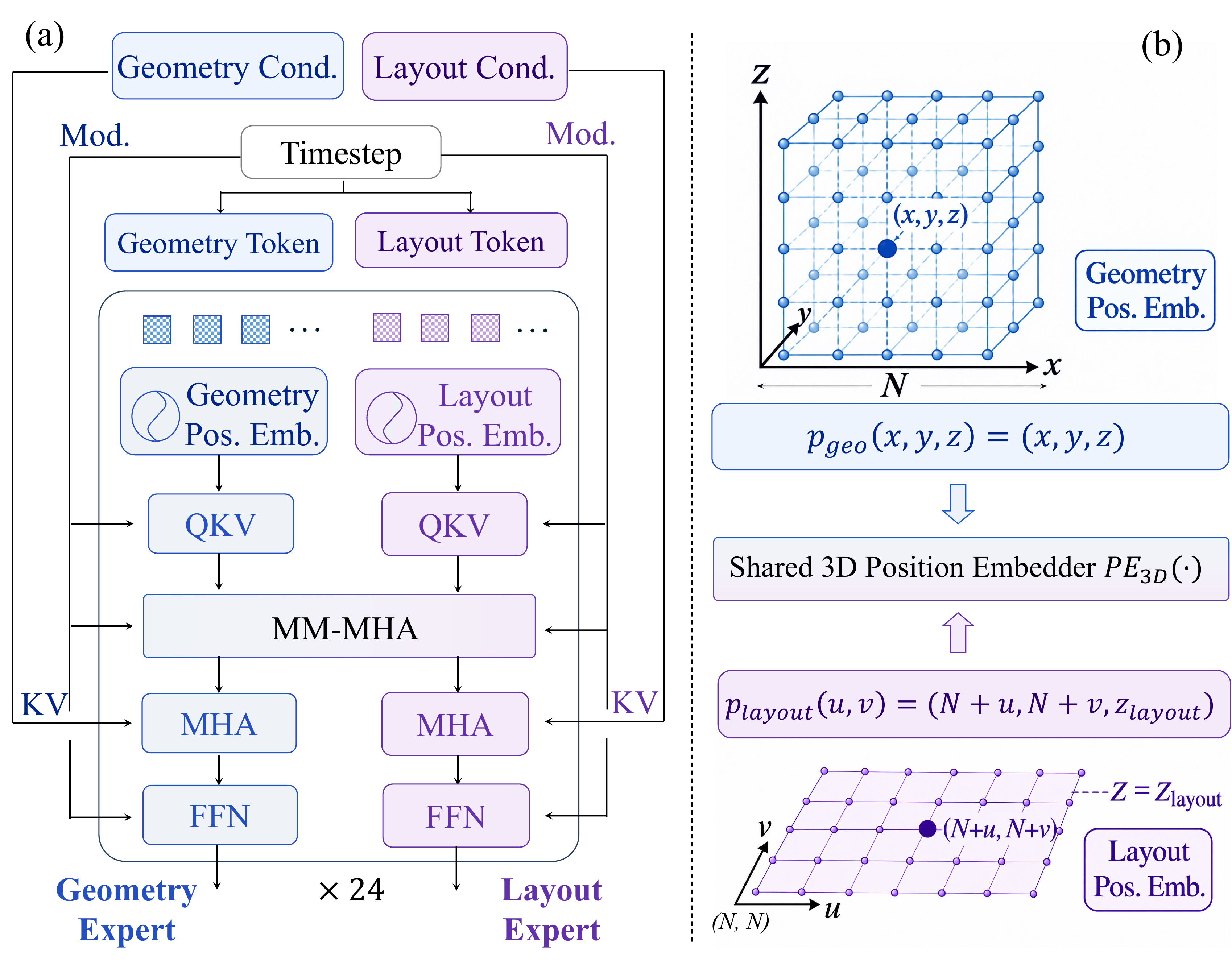}
  \caption{Geometry-layout co-generation architecture. Modality-specific experts exchange information through shared multimodal attention and use a shared 3D positional basis to improve cross-modal consistency.}
  \label{fig:network}
\end{wrapfigure}

As shown in Fig.~\ref{fig:pipeline}, our architecture adopts a Mixture-of-Transformers (MoT) design for joint geometry-layout generation. A Geometry Expert generates object geometry in a 3D voxel latent space, while a Layout Expert generates the CCM in pixel space. The two streams preserve modality-specific structure but exchange information through shared self-attention. 
Both experts are formulated as rectified flow models (see Appendix~\ref{app:network_details}).



\leavevmode\textbf{Geometry Branch.}\space
\reviewadded{%
For geometry, we represent $S_k$ as a binary voxel grid
$\{0,1\}^{N\times N\times N}$ in canonical object space.
Following recent 3D generative models~\citep{xiang2025structured},
a lightweight VAE compresses this discrete grid into a
low-resolution continuous feature grid
$G_k \in \mathbb{R}^{D\times D\times D\times C_G}$.
The noisy feature grid is serialized into a sequence of tokens and combined with 3D positional embeddings before being fed into DiT blocks for denoising.}
\par\medskip\leavevmode\textbf{Layout Branch.}\space
\reviewadded{%
 Unlike natural images with complex texture statistics, CCMs usually exhibit smooth spatial variation over visible object surfaces. 
Motivated by recent pixel-space generative
models for dense prediction tasks~\citep{xu2025pixel, li2025back},
we directly perform CCM generation in \emph{pixel space}.
The noisy CCM is tokenized by a strided convolution, combined with shared geometry-layout positional embeddings (Sec.~\ref{sec:shared_pe}), and fed into DiT blocks for denoising. 
We predict CCMs in cropped object space to preserve resolution
for small objects and avoid allocating layout tokens to
irrelevant background regions.
}

Since objects are often partially occluded, the model
must infer complete amodal shape from both local appearance
and global scene context.
\reviewadded{We condition the geometry expert on DINOv2~\citep{oquab2023dinov2} features extracted from the full image, the object mask, and the cropped object image through cross-attention. The layout expert concatenates the cropped RGB image with the noisy CCM as a local condition and incorporates full-image and mask features through cross-attention.}
Architecture and image conditioning details are provided
in Appendix~\ref{app:architecture_details}.




\par\medskip\leavevmode\textbf{Shared Geometry-Layout Position Embedding.}\space
\label{sec:shared_pe}
To facilitate the feature fusion and multi-modal self-attention, we embed both the geometry tokens and the layout tokens into a shared 3D positional space, illustrated in Fig.~\ref{fig:network}. The geometry tokens naturally distributed in the 3D space, where the 3D positional embedding can be directly applied.
The layout tokens, although arranged on a 2D image grid, are treated as points on a designated 3D plane $z=z_{\mathrm{layout}}$ with an offset $(N,N)$. This shared embedding does not assume that image-plane positions coincide with 3D surface locations; instead, it gives the two streams a common positional basis for cross-modal attention while preserving modality-specific tokenization. \reviewadded{The shared position embedding improves joint geometry-layout modeling, as shown in} (Tab.~\ref{tab:architecture_ablation}).

\par\medskip\leavevmode\textbf{Training Objective.}\space
The training objective consists of two rectified-flow losses, one for the geometry latent grid $G_k$ and one for the CCM $C_k$. Denote the patchified token sequences of $G_k$ and $C_k$ as $\bm{g}$ and $\bm{c}$, respectively. For a given timestep $t$, we sample Gaussian noise $\epsilon_g$ and $\epsilon_c$ for the two modalities and optimize
\begin{equation}
    \mathcal{L} = \lambda_1 \mathcal{L}_{CFM}^{t, \epsilon_g}(\bm{g}) + \lambda_2 \mathcal{L}_{CFM}^{t, \epsilon_c}(\bm{c}).
\end{equation}
In practice, we set $\lambda_1 = \lambda_2 = 1$.


\subsection{Scene Assembly}
\label{sec:scene_assemble}

\reviewadded{Using the dense correspondences defined in Sec.~\ref{sec:dense_corr}, we estimate each object's similarity transformation by minimizing}
\begin{equation}
s^{*},R^{*},t^{*}
=
\arg\min_{s,R,t}
\sum_{i\in\Omega}
\left\|
P_i-(sRC_i+t)
\right\|_2^2,
\quad R\in SO(3),\;s>0.
\end{equation}
\reviewadded{Here, $C_i$ and $P_i$ denote the canonical coordinate from the CCM and the corresponding scene point from the PCM at pixel $i$, respectively, and $\Omega$ contains pixels where the object mask, CCM prediction, and PCM are all valid.}

To reduce the influence of noisy CCM predictions and depth
outliers, we run RANSAC~\citep{fischler1981random} over the dense
correspondences and solve the final alignment on the inlier
set using the closed-form Umeyama algorithm~\citep{umeyama1991least}.
\reviewadded{Filtering and robust-estimation details are provided in Appendix~\ref{app:alignment}.}
\reviewadded{Applying the recovered transformation to the generated canonical geometry $S_k$ yields:}


\begin{center}
{\mbox{%
$S_{\mathrm{scene}}
=
\left\{
s_k^{*}R_k^{*}x+t_k^{*}
\mid x\in S_k
\right\}.$%
}}
\end{center}


\subsection{Training Pipeline}
\label{sec:training_pipeline}
\begin{figure}[htbp]
\centering
  \includegraphics[width=0.9\linewidth,height=0.72\textheight,keepaspectratio]{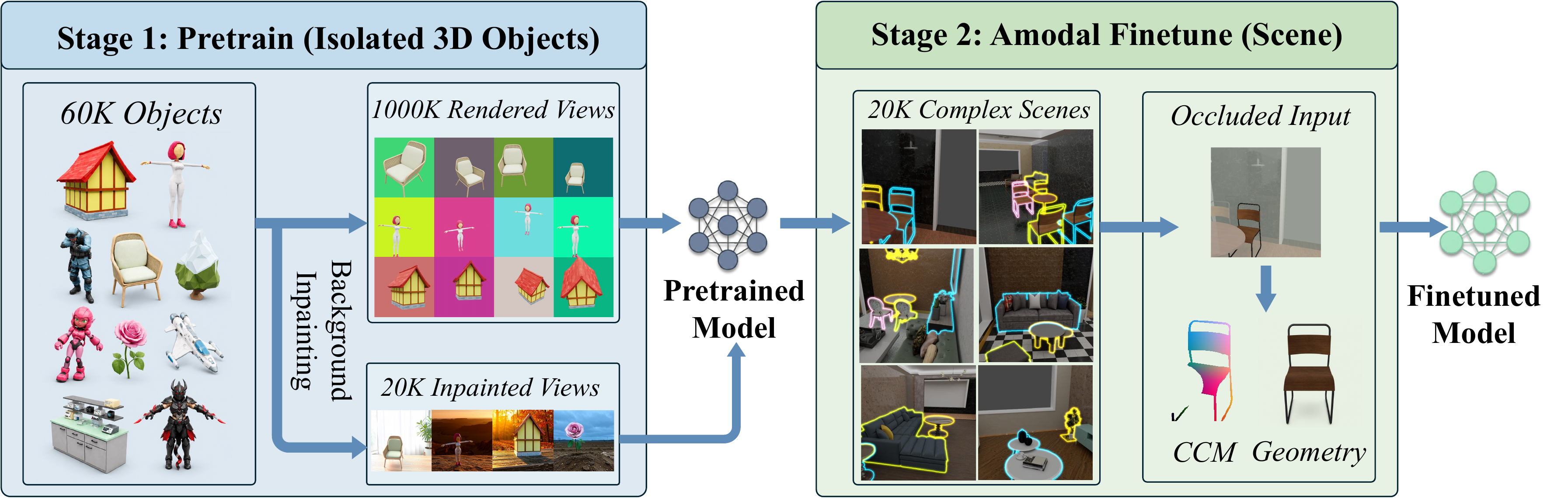}
  \caption{\reviewadded{Two-stage training pipeline. Object-level pre-training learns geometry and CCM priors from isolated-object renderings and background-completed views; scene-level fine-tuning adapts the model to occluded inputs for amodal object reconstruction.}}
  \label{fig:training}
\end{figure}

\reviewadded{Our representation enables a two-stage training pipeline, illustrated in Fig.~\ref{fig:training}.}

\noindent\textbf{Pre-training.}
\reviewadded{We pre-train on isolated 3D object assets~\citep{deitke2023objaverse}, selecting 60K objects and rendering 1M object-centric views. Since CCM is defined in canonical object space, these renderings provide direct supervision for both canonical geometry and CCMs without requiring scene-level layout annotations. To reduce the appearance gap between isolated renders and real scene images, we also use 20K photo-realistic object views with generated backgrounds (Appendix~\ref{app:data_construction}).}

\noindent\textbf{Fine-tuning.}
\reviewadded{The pretrained model can generate accurate geometry and CCMs for simple, mostly unoccluded object observations. Real scene images, however, often contain partial visibility, mutual occlusion, and diverse camera viewpoints, requiring amodal object reasoning. We therefore fine-tune the model on 20K 3D-FRONT~\citep{fu20213d} scene views by sampling occluded object instances, adapting the object-level prior to scene-level inputs.}

\section{Experiment}


\subsection{Experimental Setup}\label{sec:experimental_setup}
\noindent{Our implementation follows the two-stage training pipeline in Sec.~\ref{sec:training_pipeline}. Training settings and inference details are provided in Appendix~\ref{app:implementation}.}

\noindent{\textbf{Benchmarks.} Following standard convention, we use 3D-Future Scene~\citep{fu2021future} benchmark, containing only common indoor scenes;
and BlendSwap~\citep{blendswap} benchmark, which covers indoor and outdoor environments as well as realistic and cartoon-style appearances, providing a better demonstration of cross-domain generalization capability. 
\reviewadded{Furthermore, we qualitatively evaluate our method on a large number of in-the-wild inputs, including real, photorealistic, and stylized images, with results shown in the video and Appendix~\ref{app:scene_comparison}.}}

\noindent{\textbf{Baselines.} We compare with SOTA image-based scene generation methods, i.e. Gen3DSR~\citep{ardelean2025gen3dsr}, MIDI~\citep{huang2025midi}, SceneGen~\citep{meng2025scenegen}, and SAM3D~\citep{chen2026sam}.
All methods receive the same scene RGB image and instance masks, factoring out segmentation quality. Unless otherwise specified, all quantitative results use the same normalization protocol and alignment procedure across methods.}

\noindent{\textbf{Metrics.} We evaluate object geometry using Chamfer Distance (CD), F-score (FS) with threshold $\tau=0.1$, and Earth Mover's Distance (EMD). For scene layout, we follow SAM3D and report 3D-IoU, ICP-Rot, 2D-IoU, and ADD-S. Dataset details, metric definitions, and normalization and alignment protocols are provided in Appendix~\ref{app:evaluation_protocol}.}

\begin{table}[t]
\reviewtableclean
\setlength{\abovecaptionskip}{0pt}
\setlength{\belowcaptionskip}{6pt}
\caption{Quantitative comparison on BlendSwap and 3D-Future Scene. Note that \emph{object geometry} is evaluated using CD, FS@0.1, and EMD, while \emph{scene layout} is evaluated using 3D-IoU, ICP-Rot, 2D-IoU, and ADD-S. Best and second-best results are highlighted.}
\label{tab:all_comparison}
\centering
\fontsize{7}{8.5}\selectfont
\setlength{\tabcolsep}{1.0pt}
\begin{tabular}{@{}lccc|cccc@{\hspace{7pt}}ccc|cccc@{}}
\toprule
\multirow{2}{*}{\textbf{Method}} & \multicolumn{7}{c}{\textbf{BlendSwap}} & \multicolumn{7}{c}{\textbf{3D-Future Scene}} \\
\cmidrule(lr){2-8} \cmidrule(lr){9-15}
& CD$\downarrow$ & FS@0.1$\uparrow$ & \multicolumn{1}{c}{EMD$\downarrow$} & 3D-IoU$\uparrow$ & ICP-Rot$\downarrow$ & 2D-IoU$\uparrow$ & ADD-S$\downarrow$
& CD$\downarrow$ & FS@0.1$\uparrow$ & \multicolumn{1}{c}{EMD$\downarrow$} & 3D-IoU$\uparrow$ & ICP-Rot$\downarrow$ & 2D-IoU$\uparrow$ & ADD-S$\downarrow$ \\
\midrule
Gen3DSR & 0.072 & 0.582 & 0.263 & 0.359 & 15.54 & 0.608 & 0.163 & 0.067 & 0.576 & 0.263 & 0.502 & 11.37 & 0.638 & 0.107 \\
MIDI & 0.036 & 0.768 & 0.230 & 0.229 & 9.366 & 0.472 & 0.449 & 0.039 & 0.761 & 0.249 & 0.280 & 14.53 & 0.387 & 0.164 \\
SceneGen & 0.032 & 0.794 & 0.238 & 0.185 & 15.08 & 0.394 & 0.591 & 0.025 & 0.823 & 0.253 & 0.446 & 17.63 & 0.493 & 0.094 \\
SAM3D & \cellcolor{second}{0.027} & \cellcolor{second}{0.817} & \cellcolor{best}{0.163} & \cellcolor{second}{0.520} & \cellcolor{second}{7.566} & \cellcolor{second}{0.672} & \cellcolor{second}{0.078} & \cellcolor{best}{0.014} & \cellcolor{best}{0.866} & \cellcolor{best}{0.169} & \cellcolor{second}{0.596} & \cellcolor{second}{6.272} & \cellcolor{second}{0.639} & \cellcolor{second}{0.085} \\
Ours & \cellcolor{best}{{0.021}} & \cellcolor{best}{{0.843}} & \cellcolor{second}{{0.169}} & \cellcolor{best}{{0.727}} & \cellcolor{best}{{5.616}} & \cellcolor{best}{{0.783}} & \cellcolor{best}{{0.031}} & \cellcolor{second}{0.015} & \cellcolor{second}{0.845} & \cellcolor{second}{0.176} & \cellcolor{best}{0.694} & \cellcolor{best}{5.485} & \cellcolor{best}{0.729} & \cellcolor{best}{0.064} \\
\bottomrule
\end{tabular}
\end{table}

\subsection{Scene Generation Results}
We first evaluate full compositional scene reconstruction, including both object-level geometry and scene-level layout, to demonstrate the significance of dense correspondence-based layout.

\paragraph{Quantitative comparison.}
Tab.~\ref{tab:all_comparison} reports quantitative results on both object geometry and scene layout. For \emph{object geometry}, Mira-Scene achieves competitive performance, even with slightly better CD and F-score than SAM3D on BlendSwap. Note that SAM3D benefits from a much larger-scale data engine and substantially more object-level training data, whereas Mira-Scene is trained with only 60K open-source object assets. 
\reviewadded{Mira-Scene's advantage is most pronounced in \emph{scene layout}. It improves 3D-IoU by 39.8\% on BlendSwap and 16.4\% on 3D-Future Scene relative to SAM3D, the strongest baseline on this metric. These gains support dense CCM--PCM correspondence as a more learnable and generalizable layout representation than sparse pose regression.}

\begin{figure}[t!]
\centering
  \includegraphics[width=0.88\linewidth,height=0.72\textheight,keepaspectratio]{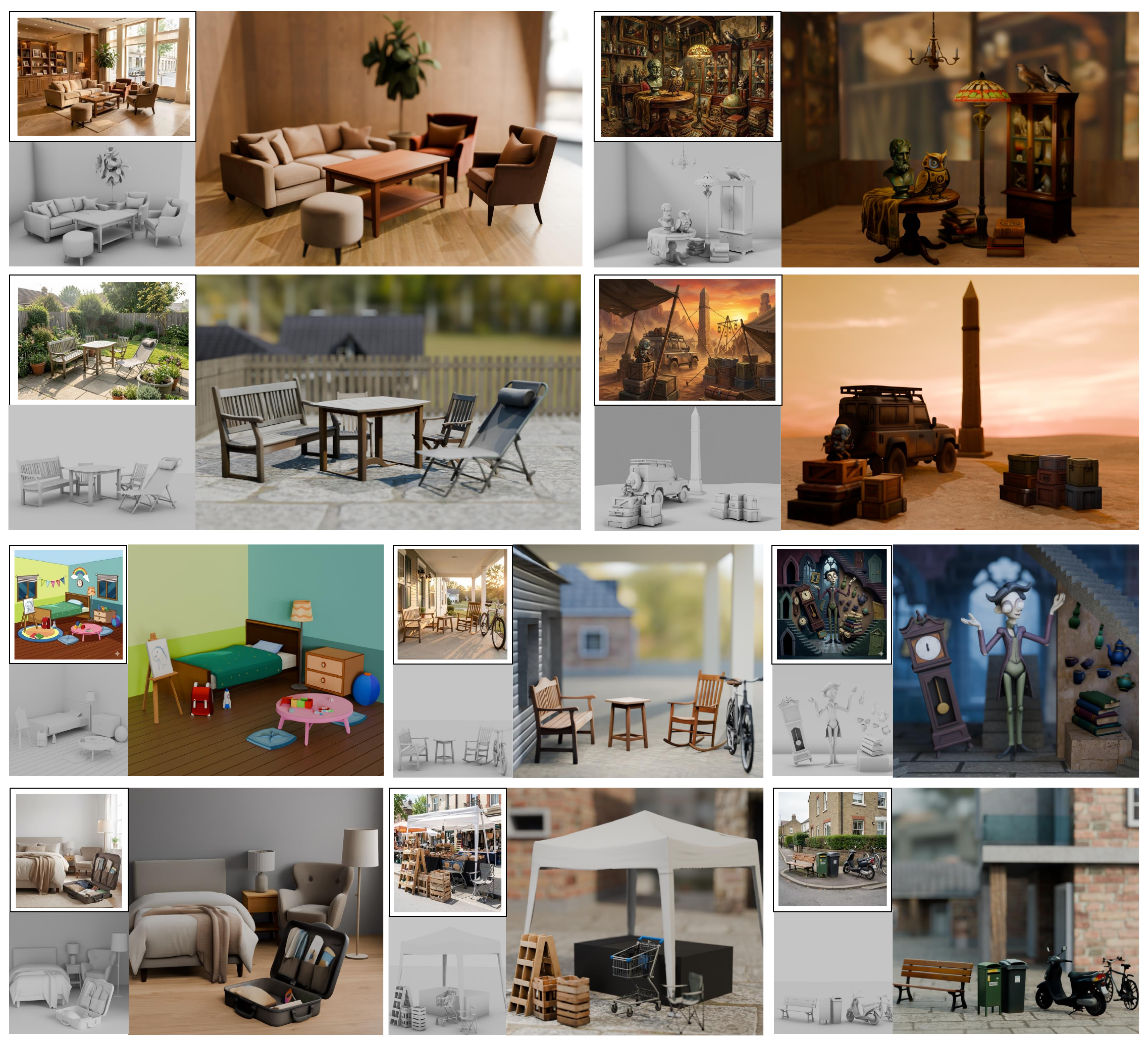}
  \caption{Compositional scene reconstruction results produced by Mira-Scene. Given a single image, our method generates object-level 3D assets and places them into a coherent scene layout.}
  \label{fig:result_show}
\end{figure}

\paragraph{Qualitative comparison.}
\noindent{Fig.~\ref{fig:result_show} presents out-of-domain scene reconstructions produced by Mira-Scene.}
\reviewadded{The comparisons with state-of-the-art methods are provided in Fig.~\ref{fig:comparison} (Appendix~\ref{app:scene_comparison}).}

Across indoor, outdoor, realistic, and stylized examples, the compared methods exhibit different failure modes. SceneGen produces plausible results on indoor scenes close to its training distribution, but its layouts degrade on outdoor and stylized inputs. MIDI benefits from object-level 3D priors but still lacks a reliable mechanism for precise object placement. SAM3D recovers coarse layouts in many cases, but its sparse layout representation often leaves visible misalignment. 
Another recently open-sourced method, SceneMaker~\citep{shi2025scenemaker}, with a sparse layout representation, also encounters a similar problem, as shown in Appendix~\ref{app:scenemaker_comparison}.
In contrast, Mira-Scene maintains better projection consistency across views, \reviewadded{reflecting the benefit of dense CCM--PCM alignment.}
\par
\subsection{3D--2D Correspondence Analysis}\label{sec:correspondence_analysis}
Mira-Scene relies on a dense visible-surface correspondence between image pixels and canonical object coordinates. We therefore compare our formulation with CUPID~\citep{huang2025cupid}, a recent method that also models 3D--2D correspondence for image-to-3D generation.

\begin{figure}[t!]
\centering
\begin{minipage}[t]{0.46\textwidth}
\reviewtableclean
\vspace{0pt}
\setlength{\abovecaptionskip}{0pt}
\setlength{\belowcaptionskip}{6pt}
\captionof{table}{\reviewadded{3D--2D correspondence quality.} \reviewadded{CD and F-scores measure alignment between correspondence-induced and target rendered-depth point clouds.}}
\label{tab:3d-2d-correspondence}
\centering
\fontsize{7.5}{9}\selectfont
\setlength{\tabcolsep}{1.6pt}
\renewcommand{\arraystretch}{1.57}
\begin{tabular}{@{}lcccc@{}}
    \toprule
        \textbf{Method} & {2D-IoU$\uparrow$} & {CD$\downarrow$} & {FS@0.01$\uparrow$} & {FS@0.05}$\uparrow$ \\
        \midrule
        {CUPID-(Mesh+GT)}
        & {0.802} & {0.047} & {0.414} & {0.727} \\
        {Ours-(Mesh+GT)}
        & {0.795} & {0.023} & {0.456} & {0.885} \\
        {Ours-(CCM+GT)}
        & {\xmark} & {0.021} & {0.523} & {0.898} \\
        {Ours-(CCM+Mesh)}
        & {\xmark} & \cellcolor{best}{0.012} & \cellcolor{best}{0.546} & \cellcolor{best}{0.977} \\
    \bottomrule
    \end{tabular}

\end{minipage}\hfill
\begin{minipage}[t]{0.52\textwidth}
\vspace{0pt}
\centering
\includegraphics[width=\linewidth]{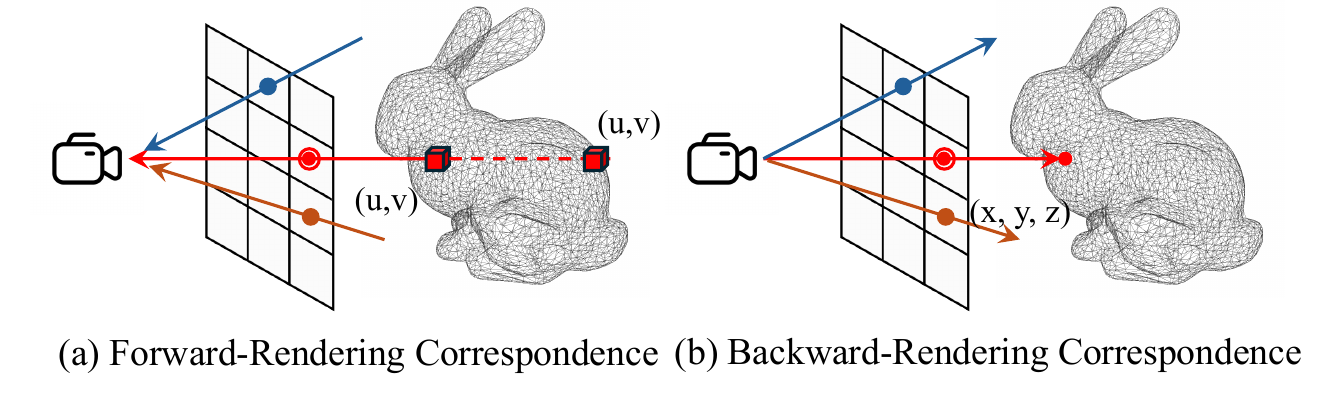}
\par
\setlength{\abovecaptionskip}{\baselineskip}
\setlength{\belowcaptionskip}{0pt}
\caption{Forward versus reverse 3D--2D correspondence. CUPID predicts where each 3D point projects in the image, which is many-to-one along camera rays. Our CCM predicts the reverse visible-pixel-to-canonical-surface mapping, which directly supports alignment with the scene-space PCM.}
\label{fig:3d-2d-correspondence}
\end{minipage}
\end{figure}

\reviewadded{As illustrated in Fig.~\ref{fig:3d-2d-correspondence},
CUPID follows a forward-rendering-like formulation: it stores, for each 3D voxel center, the pixel coordinate of its projected point in the 2D image. In contrast, Mira-Scene resembles a reverse-rendering process, \reviewadded{\reviewadded{directly recording the 3D coordinate for each 2D pixel.}}}


Although CUPID can recover the camera pose via a Perspective-n-Point (PnP) solver and align a 3D object with a 2D image, it is less suited for scene generation, which requires accurate 2D-pixel-to-3D-point correspondences to associate image observations with the point-cloud map (or scene coordinates). As shown in Fig.~\ref{fig:3d-2d-correspondence}(a), CUPID primarily models the projection from complete 3D voxels to 2D pixels; reversing this correspondence leads to a one-to-many mapping, as multiple 3D points along a camera ray can correspond to the same pixel. In contrast, Mira-Scene directly models a one-to-one mapping from 2D pixels to 3D points, \reviewadded{reducing correspondence ambiguity.}


Moreover, learning 3D-2D correspondence in 3D space is intrinsically harder than 2D. This stems from the fact that 2D semantic features (e.g., DINO) already implicitly contain the geometric information of pixel patches. As a result, predicting the 3D coordinates of a 2D token from these features is far easier than predicting the 2D coordinates of a 3D token from the same features. As shown in the first row of Fig.~\ref{fig:3d-2d-correspondence-res}, when generalizing to a new case, CUPID predicts an erroneous downward-looking camera and the corresponding geometry.



Tab.~\ref{tab:3d-2d-correspondence} reports quantitative correspondence following CUPID protocol (refer to Appendix~\ref{app:correspondence_protocol} for details). 
Although CUPID achieves slightly higher 2D mask IoU, Mira-Scene obtains substantially better CD and F-scores in the Mesh+GT setting, indicating more accurate geometric correspondence. The strong result of Ours-(CCM+Mesh) shows that the CCM predicted by our layout branch is highly consistent with the mesh produced by our geometry branch. This should be interpreted as internal consistency between the two generated outputs rather than higher fidelity than the ground-truth mesh, demonstrating the effectiveness of our joint generation framework. The visualization in the last column of Fig.~\ref{fig:3d-2d-correspondence-res} further supports this observation.

\subsection{Ablation}

\begin{wrapfigure}{R}{0.43\textwidth}
\setlength{\abovecaptionskip}{5pt}
\centering
\includegraphics[width=\linewidth]{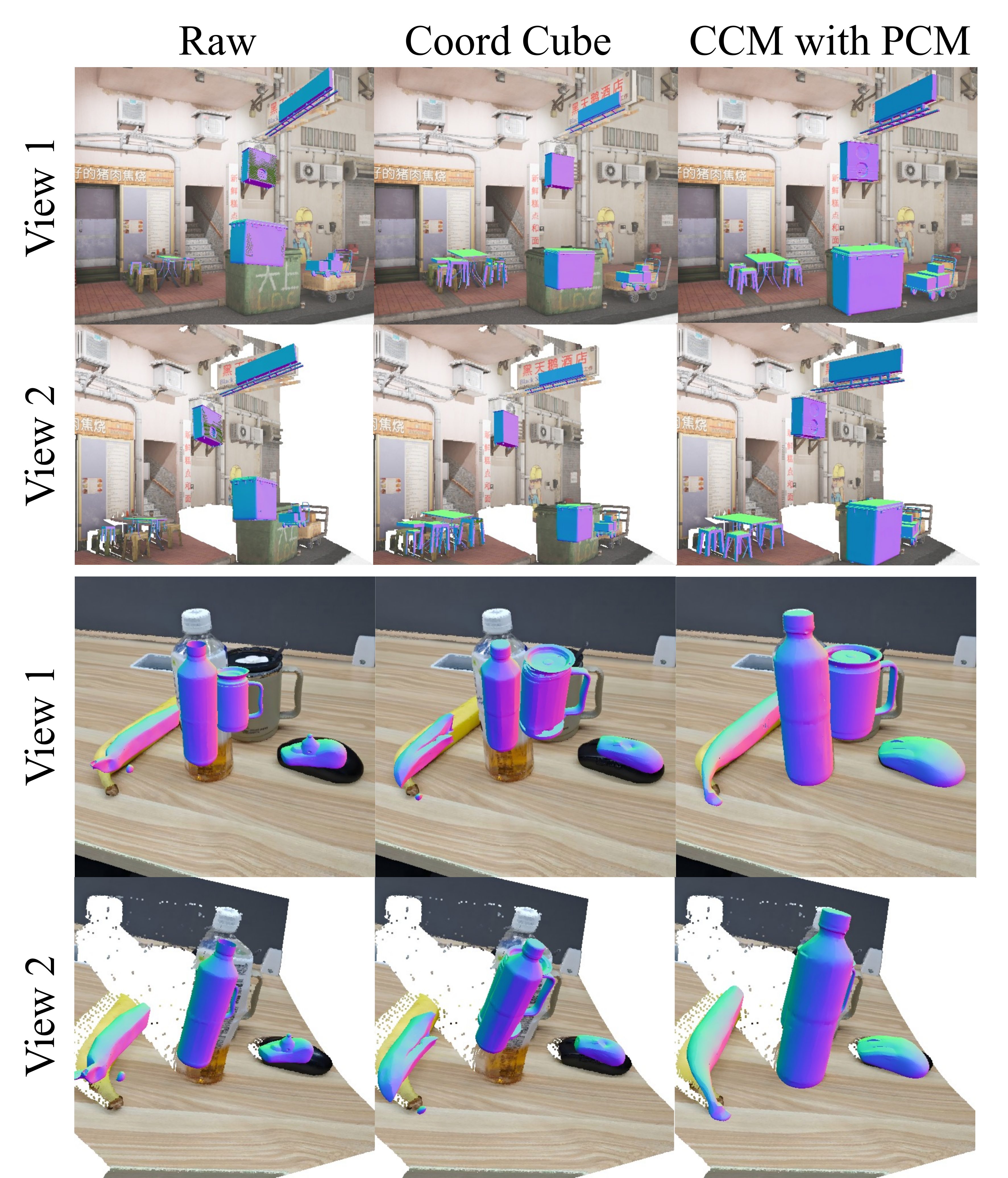}
\caption{Layout comparison. CCM with PCM improves object placement over Raw and Coord Cube.}
\label{fig:layout_comparison}
\end{wrapfigure}
\leavevmode\textbf{Layout Representation.} We first ablate the layout representation to verify whether the improvement comes from the proposed dense canonical correspondence rather than the shared architecture alone. We compare the three representations introduced in Fig.~\ref{fig:layout_representation}: \emph{Raw}, \emph{Coord Cube}, and \emph{CCM with PCM}. All variants use the same geometry-layout co-generation architecture.
\reviewadded{For fairness, Raw and Coord Cube receive monocular geometry features~\citep{wang2025moge}, giving them access to geometric priors comparable to the PCM used by CCM. Tokenization, training data, and conditioning details are provided in Appendix~\ref{sec:supp_layout_rep}.}

Tab.~\ref{tab:layout_comparison} shows that Coord Cube only slightly improves over Raw, suggesting that densifying scene-space prediction alone does not resolve the difficulty of learning unbounded scene-space targets. In contrast, CCM with PCM substantially improves layout accuracy, increasing 3D-IoU from 0.379 to \reviewadded{0.727} and 2D-IoU from 0.381 to \reviewadded{0.783} over Coord Cube. It also improves object geometry metrics, mainly because CCM is defined in canonical object space and can be trained with scalable object-level data, while Raw and Coord Cube depend more heavily on limited scene-space supervision. The visual comparison in Fig.~\ref{fig:layout_comparison} further shows that Raw and Coord Cube suffer from object drift and scale errors, while CCM produces more stable object placement. \reviewadded{CCM with PCM also outperforms Raw and Coord Cube under matched training data; the controlled comparison is provided in Appendix~\ref{app:matched_data_ablation}.}

\noindent\textbf{Network Architecture.} For the network architecture, we evaluate two variants: w/o joint attention, where the geometry and layout branches are fully decoupled without any cross-branch interaction; and w/o shared position embedding, where the layout branch uses a standard 2D positional embedding instead of the unified 3D positional encoding. For efficiency, all architecture variants are trained during pretraining on a subset of 100K images, corresponding to approximately 50K 3D objects.

Following the 3D--2D correspondence analysis above, we evaluate geometry-layout consistency by comparing the point cloud induced by predicted CCMs with the generated mesh. Tab.~\ref{tab:architecture_ablation} shows that removing joint attention causes a large degradation, reducing 2D-IoU from 0.757 to 0.535 and increasing CD from 0.017 to 0.070. This confirms that geometry and CCM should exchange information during generation. Removing shared PE leads to a smaller but consistent drop, indicating that a shared positional basis further improves cross-modal alignment. The full model achieves the best consistency across all metrics. \reviewadded{\reviewadded{See Appendix~\ref{app:ablation_visuals} for qualitative results.}}

\AddToHookNext{cmd/@makecol/before}{\setlength{\textfloatsep}{12pt}}
\begin{table}[t]
\reviewtableclean
\begin{minipage}[t]{0.50\linewidth}
\vspace{0pt}
\begin{minipage}[t]{\linewidth}
\setlength{\abovecaptionskip}{0pt}
\captionof{table}{Ablation of layout representations. CD and FS@0.1 evaluate object geometry; 3D-IoU, ICP-Rot, and 2D-IoU evaluate scene layout.}
\label{tab:layout_comparison}
\end{minipage}\par\vspace{4pt}
\centering
\fontsize{7}{8.5}\selectfont
\setlength{\tabcolsep}{2pt}
\renewcommand{\arraystretch}{1.08}
\begin{tabular}{@{}lcc@{\hspace{6pt}}ccc@{}}
    \toprule
        \textbf{Method}& CD$\downarrow$ & FS@0.1$\uparrow$
          & 3D-IoU$\uparrow$ & ICP-Rot$\downarrow$ & 2D-IoU$\uparrow$ \\
        \midrule
        {Raw}
        & {0.053} & {0.626} 
        & {0.365} & {10.12} & {0.358} \\
        Coord Cube
        & {0.049} & {0.632}
        & {0.379} & {10.03} & {0.381} \\
        CCM with PCM
        & {\reviewadded{0.021}} & {\reviewadded{0.843}}
        & {\reviewadded{0.727}} & {\reviewadded{5.616}} & {\reviewadded{0.783}} \\
    \bottomrule
    \end{tabular}

\end{minipage}\hfill
\begin{minipage}[t]{0.48\linewidth}
\vspace{0pt}
\begin{minipage}[t]{\linewidth}
\setlength{\abovecaptionskip}{0pt}
\captionof{table}{Geometry-layout co-generation ablation. Metrics follow the 3D--2D correspondence consistency evaluation in Tab.~\ref{tab:3d-2d-correspondence}.}
\label{tab:architecture_ablation}
\end{minipage}\par\vspace{4pt}
\centering
\fontsize{7}{8.5}\selectfont
\setlength{\tabcolsep}{2pt}
\renewcommand{\arraystretch}{1.08}
\begin{tabular}{@{}lcccc@{}}
    \toprule
        \textbf{Method} & 2D-IoU$\uparrow$ & CD$\downarrow$ & FS@0.01$\uparrow$ & FS@0.05$\uparrow$ \\
        \midrule
        w/o Joint Attention
        & {0.535} & {0.070} & {0.125} & {0.505} \\
        w/o Shared PE.
        & {0.747} & {0.019} & {0.349} & {0.925} \\
        {Full}
        & {0.757} & {0.017} & {0.383} & {0.940} \\
    \bottomrule
    \end{tabular}

\end{minipage}
\end{table}




\subsection{Applications}

\leavevmode\reviewadded{Our compositional reconstruction enables \reviewadded{downstream applications using object-level scene assets.} As shown in Fig.~\ref{fig:application} (Appendix~\ref{app:applications}), users can edit reconstructed scenes by removing, repositioning, reconfiguring, rigging, or animating individual objects while preserving a coherent spatial layout. \reviewadded{These object assets can also be exported to} interactive editing tools, embodied-AI simulators, and physics engines for perception, planning, manipulation, or physical simulation.}

\section{Discussion and Conclusions}
\leavevmode\reviewadded{We presented Mira-Scene, a framework for single-image compositional 3D scene reconstruction. Our formulation replaces sparse object pose regression with a dense, bounded CCM representation and recovers object placement through CCM--PCM alignment, reducing dependence on scarce scene-level 3D supervision. Using separate geometry and layout streams with shared attention and a shared positional basis, our co-generation model preserves the distinct structures of 3D geometry and pixel-aligned CCMs while improving their consistency. Experiments on indoor, outdoor, synthetic, and in-the-wild inputs show that Mira-Scene achieves competitive object geometry and significantly more accurate scene layouts than strong baselines. The ability to learn from abundant object-level 3D assets provides strong potential for further scaling. {Please refer to Appendix~\ref{app:limitations} for discussion about limitations and future work.}}

\subsection*{AI use statement}
\reviewadded{We used generative AI tools to polish the manuscript for clarity and readability and to shorten the text to meet the page limit. We also used FLUX2.0 for background completion of rendered object images used in training (Appendix~\ref{app:data_construction}), and Gemini 3 Pro Image (Nano Banana Pro) and FLUX2 to generate a subset of the input images for qualitative evaluation (Appendix~\ref{app:evaluation_datasets}). The authors take responsibility for the final content of this work, including all AI-assisted text and generated data.}

\clearpage
\bibliographystyle{iclr2027_conference}
\bibliography{references}

\clearpage
\appendix
\section*{\reviewadded{Appendix}}


\section{\reviewadded{Additional Experimental Results}}
\label{app:additional_results}
\label{app:additional_analyses}
\label{app:additional_qualitative}
\subsection{\reviewadded{Qualitative Comparison with Scene Generation Methods}}
\label{app:scene_comparison}
\begin{figure}[H]
\reviewmovedcolor
\centering
  \includegraphics[width=0.99\linewidth,height=0.72\textheight,keepaspectratio]{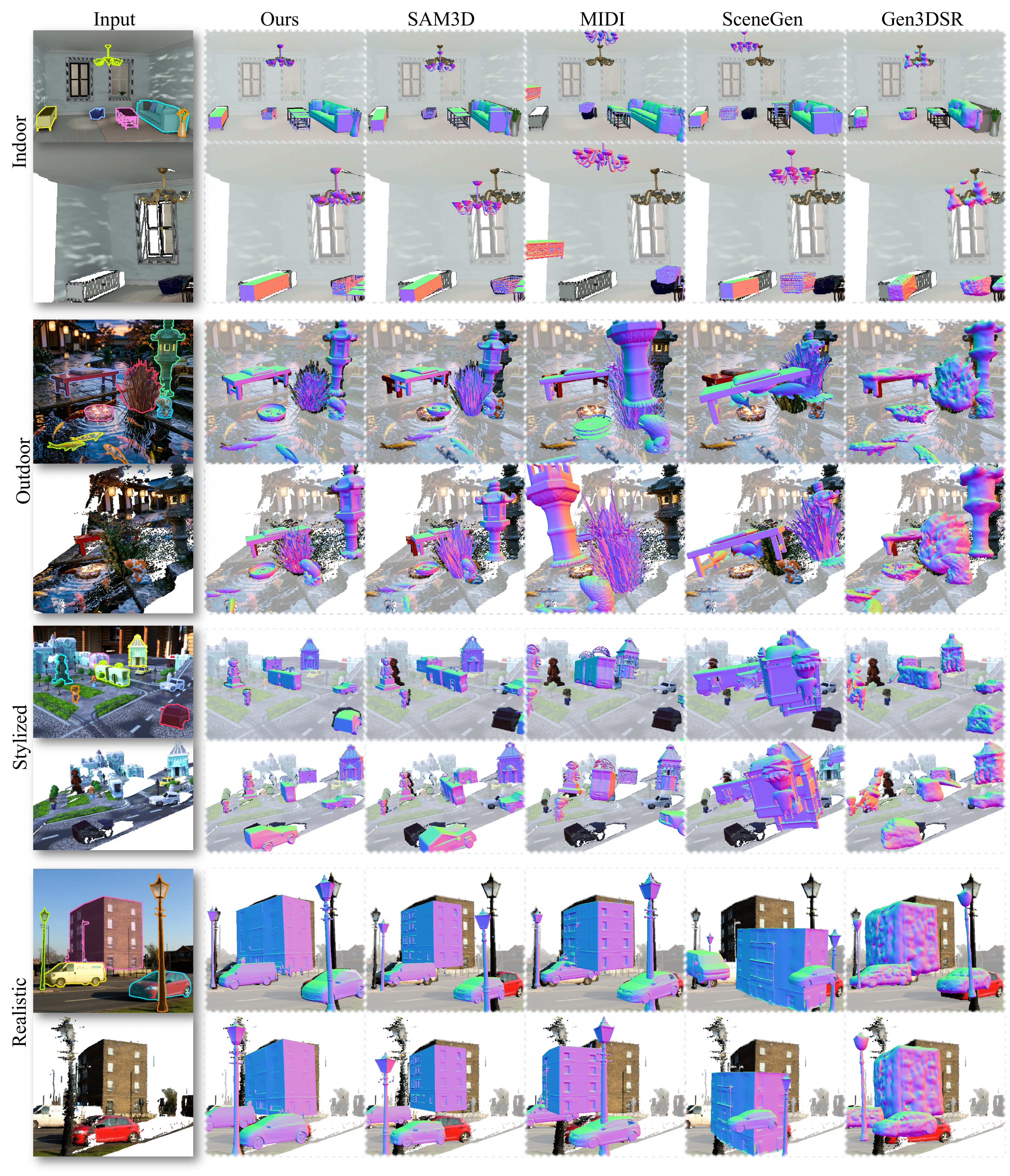}
  \caption{
Qualitative comparison with state-of-the-art single-image 3D scene generation and reconstruction methods.
From left to right: input image, Ours, SAM3D, MIDI, SceneGen, and Gen3DSR.
For each scene, we show normal renderings overlaid with the input image from the input view and an additional novel view.
Top to bottom: indoor BlendSwap scene, AI-generated outdoor image, synthetic outdoor BlendSwap scene, and real-world outdoor image.
}
  \label{fig:comparison}
\end{figure}

\begin{reviewmoved}
For comparison, we provide normal renderings overlaid with estimated scene point maps in the input view and a novel view with SOTA methods in Fig.~\ref{fig:comparison}. This visualization makes layout errors visible: misplaced objects no longer align with the image evidence in the input view and drift more clearly under novel views.
\end{reviewmoved}

\clearpage
\begingroup
\begingroup
\reviewmarksfalse
\subsection{\reviewadded{Qualitative Comparison with CUPID}}
\label{app:correspondence_visuals}
\noindent\reviewadded{Fig.~\ref{fig:3d-2d-correspondence-res} complements the correspondence analysis in Sec.~\ref{sec:correspondence_analysis}. The last column overlays CCM-induced points on our mesh to visualize geometry--correspondence consistency.}

\subsection{\reviewadded{Additional Ablation Visualizations}}
\label{app:ablation_visuals}
\noindent\reviewadded{Fig.~\ref{fig:ablation} visualizes the architecture variants evaluated in Table~\ref{tab:architecture_ablation}, with CCM-induced points overlaid on the generated meshes.}

\par
\noindent
\begin{minipage}[t]{0.48\textwidth}
\vspace{0pt}
\begingroup
\reviewmovedcolor
\centering
\includegraphics[width=0.9\linewidth]{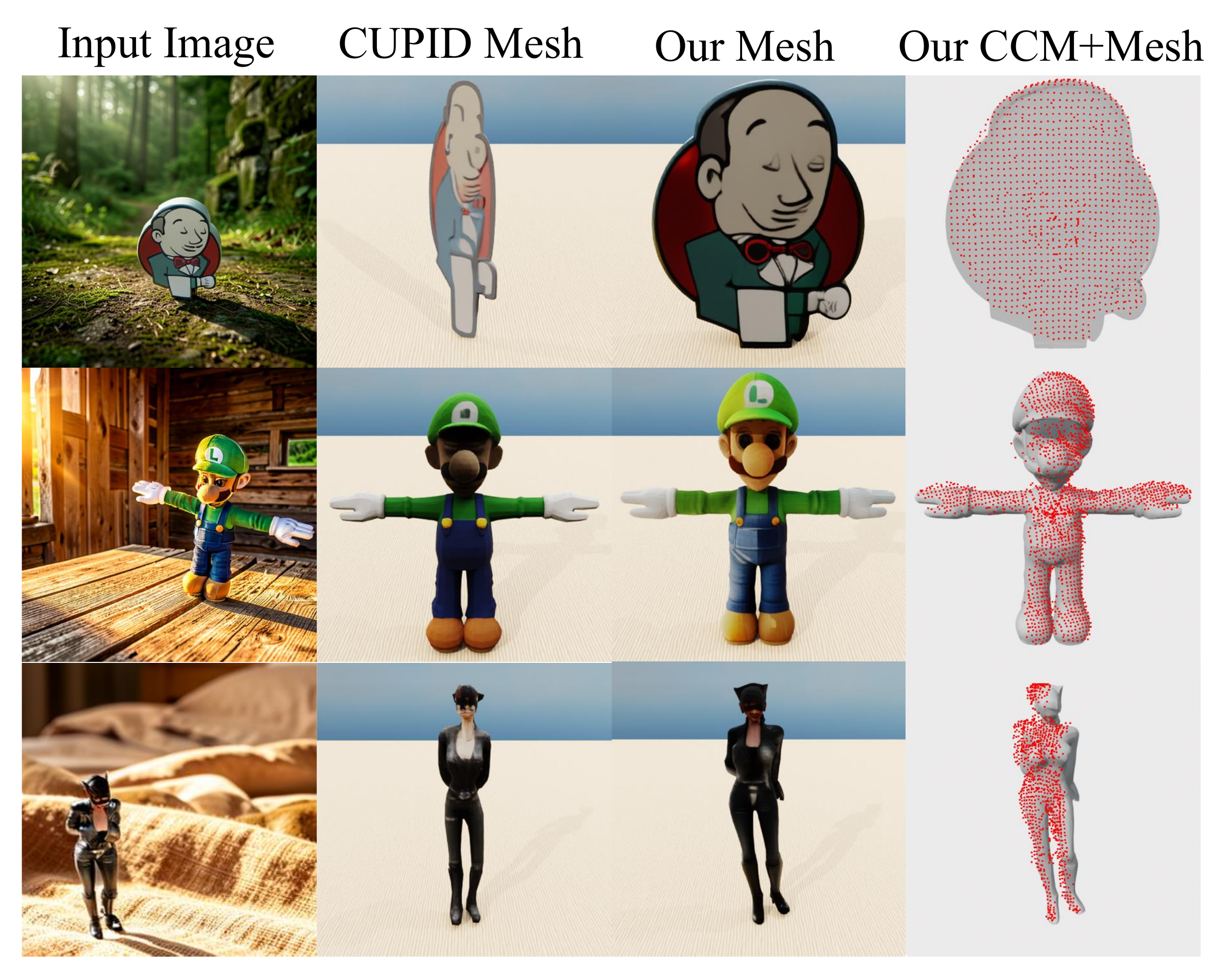}
\captionof{figure}{Compared with CUPID, Mira-Scene produces more plausible geometry and better CCM--mesh consistency.}
\label{fig:3d-2d-correspondence-res}
\par\endgroup
\end{minipage}\hfill
\begin{minipage}[t]{0.48\textwidth}
\vspace{0pt}
\begingroup
\reviewmovedcolor
\centering
\includegraphics[width=\linewidth]{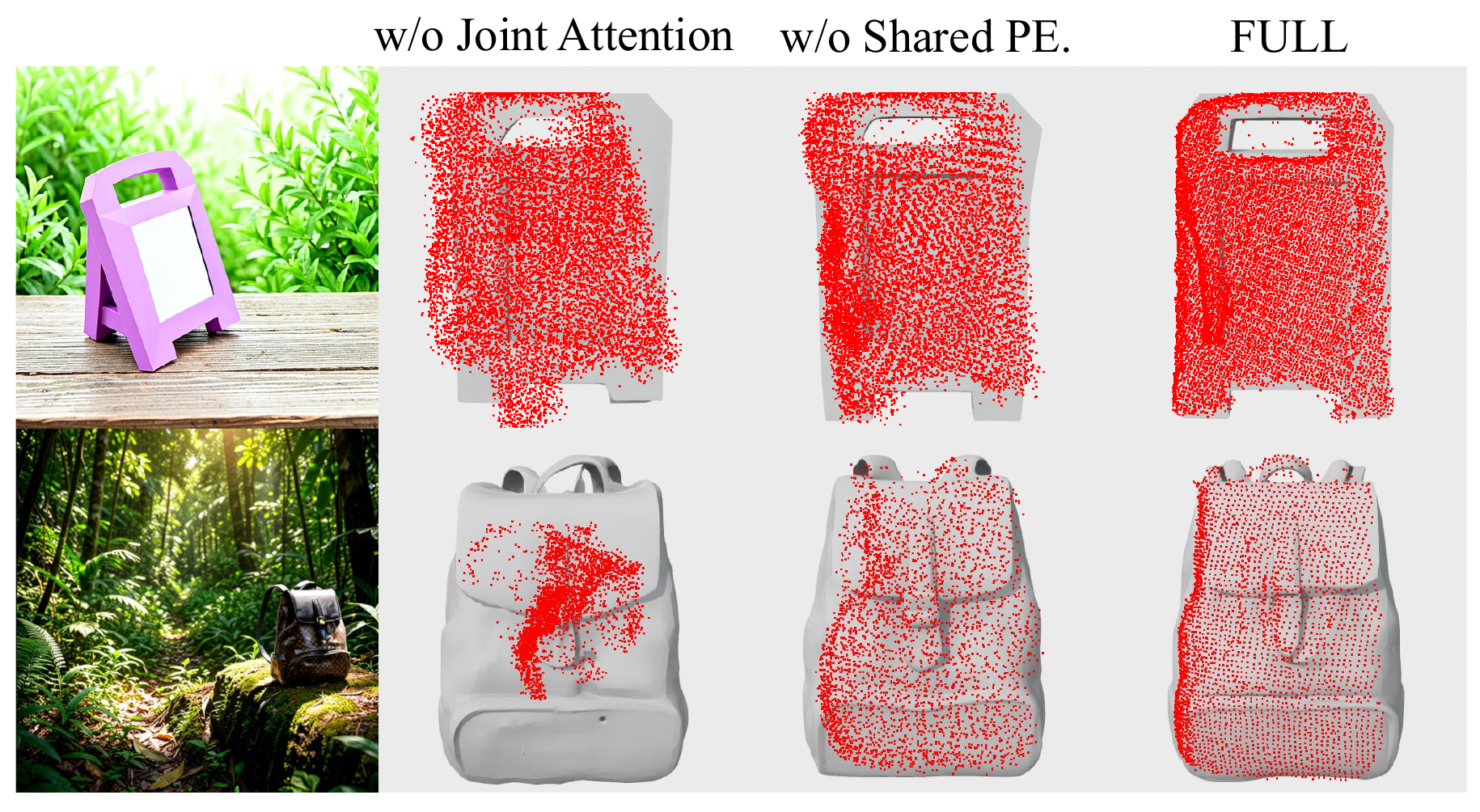}
\captionof{figure}{Architecture ablation for geometry-layout co-generation. Joint attention is important for maintaining consistency between generated geometry and predicted CCM, while shared positional encoding provides additional alignment improvement.}
\label{fig:ablation}
\par\endgroup
\end{minipage}
\par

\par
\endgroup
\subsection{\reviewadded{Layout Representation under Matched Training Data}}
\label{app:matched_data_ablation}
\begin{reviewaddedblock}
To separate representation benefits from additional object-level supervision, we train a CCM variant using the same background-completed object views and scene-level data as Raw and Coord Cube, without the additional isolated object assets.

Table~\ref{tab:matched_data_ablation} shows that CCM with PCM outperforms Raw and Coord Cube under matched training data. Additional object-level supervision further improves layout accuracy, supporting both the representation and the benefit of object-level data.
\end{reviewaddedblock}

\begingroup
\reviewmarksfalse
\subsection{\reviewadded{Qualitative Comparison with SceneMaker}}
\label{app:scenemaker_comparison}

\noindent
We compare with SceneMaker~\citep{shi2025scenemaker} on BlendSwap using at most 5 selected objects per scene, following its reported setting. As shown in Fig.~\ref{fig:scenemaker_results}, SceneMaker results show object substitution and inaccurate scale, pose, or relative layout.

\par
\endgroup
\par
\noindent
\begin{minipage}[t]{0.48\textwidth}
\vspace{0pt}
\begingroup
\reviewtableclean
\reviewaddedcolor
\centering
\setlength{\belowcaptionskip}{6pt}
\captionof{table}{Layout representation ablation under matched training data.}
\label{tab:matched_data_ablation}
\setlength{\tabcolsep}{3pt}
\begin{tabularx}{\linewidth}{@{}Xcc@{}}
\toprule
Method & 3D-IoU$\uparrow$ & 2D-IoU$\uparrow$ \\
\midrule
Raw & 0.365 & 0.358 \\
Coord Cube & 0.379 & 0.381 \\
CCM with PCM (matched data) & 0.537 & 0.662 \\
CCM with PCM (full) & 0.727 & 0.783 \\
\bottomrule
\end{tabularx}

\par\endgroup
\end{minipage}\hfill
\begin{minipage}[t]{0.48\textwidth}
\vspace{0pt}
\begingroup
\centering
\reviewmarksfalse
    \includegraphics[width=0.80\linewidth]{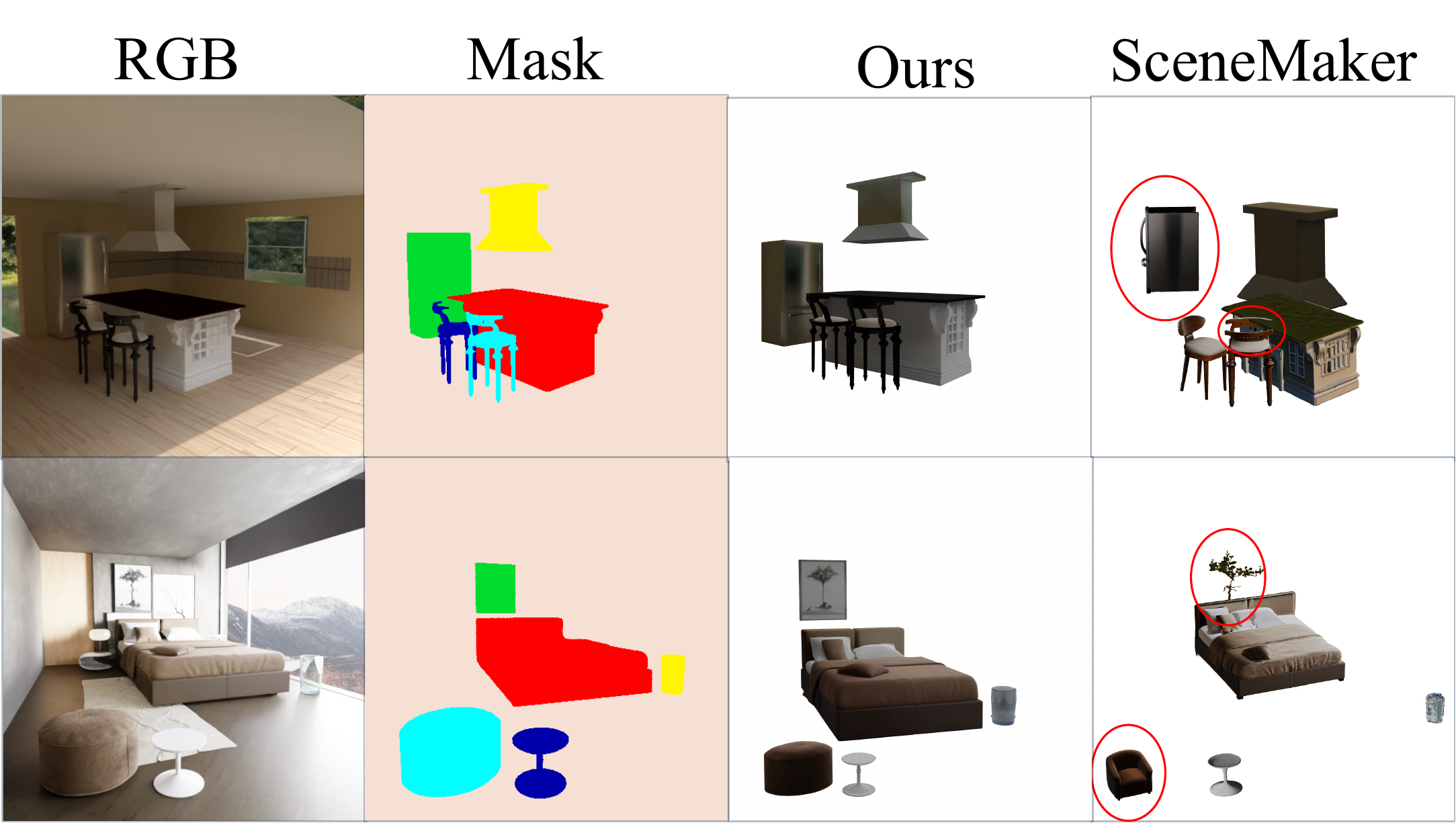}
    \captionof{figure}{\reviewreplace{Qualitative comparison with SceneMaker~\mbox{\citep{shi2025scenemaker}} on BlendSwap. Red circles mark representative object substitution and pose errors in the SceneMaker results.}{Qualitative comparison with SceneMaker. Red circles mark representative object substitution and pose errors.}}
    \label{fig:scenemaker_results}
\par\endgroup
\end{minipage}
\par

\par

\Needspace{10\baselineskip}
\subsection{\reviewadded{Additional Visual Quality Evaluation}}
\label{app:object_visual_quality}
\begin{reviewaddedblock}
To complement the geometry and layout evaluation in the main paper, we additionally evaluate object visual quality using PSNR. We compute PSNR using the evaluation implementation of SceneGen~\citep{meng2025scenegen}, with RGB values normalized to $[0,1]$. As shown in Table~\ref{tab:object_visual_quality}, Mira-Scene achieves a slightly higher PSNR than SAM3D~\citep{chen2026sam} (18.8 vs.\ 18.6). We include the previously reported CD and 3D-IoU results, together with training data information, for context.
\end{reviewaddedblock}

\begin{table}[H]
\reviewtableclean
\reviewaddedcolor
\centering
\setlength{\belowcaptionskip}{6pt}
\caption{Additional visual quality evaluation using PSNR. CD and 3D-IoU are reproduced from Table~\ref{tab:all_comparison} for reference.}
\label{tab:object_visual_quality}
\small
\setlength{\tabcolsep}{4pt}
\begin{tabular}{ll l ccc}
\toprule
Method & 3D data scale & Data preparation & CD$\downarrow$ & PSNR$\uparrow$ & 3D-IoU$\uparrow$ \\
\midrule
SAM3D & Million+ & Private data & 0.027 & 18.6 & 0.520 \\
Ours & 60K & Public data, automatic pipeline & 0.021 & 18.8 & 0.727 \\
\bottomrule
\end{tabular}
\end{table}

\subsection{\reviewadded{Correspondence Errors and Alignment Robustness}}
\label{app:alignment_robustness}
\begin{reviewaddedblock}
We analyze correspondence errors and the choice of downstream alignment solver on the BlendSwap benchmark. For valid object-mask pixels, correspondence error measures the Euclidean distance between each CCM point transformed into scene space by the estimated similarity transform and its corresponding PCM point. Table~\ref{tab:correspondence_error_distribution} shows a long-tailed error distribution: the median error is 0.046, while the 95th and 99th percentiles reach 0.395 and 0.882. This motivates robust alignment to reduce the influence of outliers.
\end{reviewaddedblock}

\begin{table}[H]
\reviewtableclean
\reviewaddedcolor
\centering
\setlength{\belowcaptionskip}{6pt}
\caption{Distribution of predicted correspondence errors on the BlendSwap benchmark.}
\label{tab:correspondence_error_distribution}
\small
\begin{tabular}{lrrrrrr}
\toprule
Statistic & Median & Mean & p75 & p90 & p95 & p99 \\
\midrule
Error & 0.046 & 0.097 & 0.103 & 0.205 & 0.395 & 0.882 \\
\bottomrule
\end{tabular}
\end{table}

\begin{reviewaddedblock}
We further ablate the choice of geometric alignment solver. RANSAC+Umeyama improves 3D-IoU by approximately 13.4\% relative to Umeyama without RANSAC. Iteratively reweighted least squares (IRLS)~\citep{holland1977robust} with a Huber loss~\citep{huber1992robust} performs similarly, yielding an additional 1.0\% relative improvement over RANSAC+Umeyama. We therefore retain RANSAC+Umeyama for its simplicity and lack of additional training; implementation details are provided in Appendix~\ref{app:alignment}.
\end{reviewaddedblock}

\subsection{\reviewadded{Layout Accuracy under Occlusion}}
\label{app:occlusion_analysis}
\begin{reviewaddedblock}
Our scene-level fine-tuning uses 3D-FRONT views to pair partial image observations with complete 3D objects, supporting amodal object reconstruction and layout recovery under occlusion. Table~\ref{tab:occlusion_analysis} reports layout accuracy for objects grouped by occlusion rate. The 3D-IoU decreases from 0.752 for mostly visible objects to 0.635 for heavily occluded objects, showing the increased difficulty of layout recovery under severe occlusion while retaining substantial overlap with the ground-truth object boxes.
\end{reviewaddedblock}

\begin{table}[H]
\reviewtableclean
\reviewaddedcolor
\centering
\setlength{\belowcaptionskip}{6pt}
\caption{Layout accuracy under different occlusion levels. Object proportion denotes the share of evaluated objects in each group.}
\label{tab:occlusion_analysis}
\small
\begin{tabular}{lcc}
\toprule
Occlusion rate & Object proportion & 3D-IoU$\uparrow$ \\
\midrule
Below 0.2 (mostly visible) & 41.8\% & 0.752 \\
$[0.2,0.6]$ & 35.5\% & 0.724 \\
Above 0.6 (heavily occluded) & 22.7\% & 0.635 \\
\bottomrule
\end{tabular}
\end{table}

\Needspace{0.88\textheight}
\subsection{\reviewadded{Additional Scene Reconstructions}}
This section provides additional qualitative results on diverse inputs, including indoor, outdoor, real, and stylized scenes. For each example, we show the input image, object decomposition, reconstructed compositional scene, and novel-view renderings. These results complement the quantitative comparison in the main paper and further illustrate the robustness of Mira-Scene under varied scene layouts and visual styles.

\begin{figure}[H]
\centering
  \centering
  \includegraphics[width=0.95\textwidth,height=0.72\textheight,keepaspectratio]{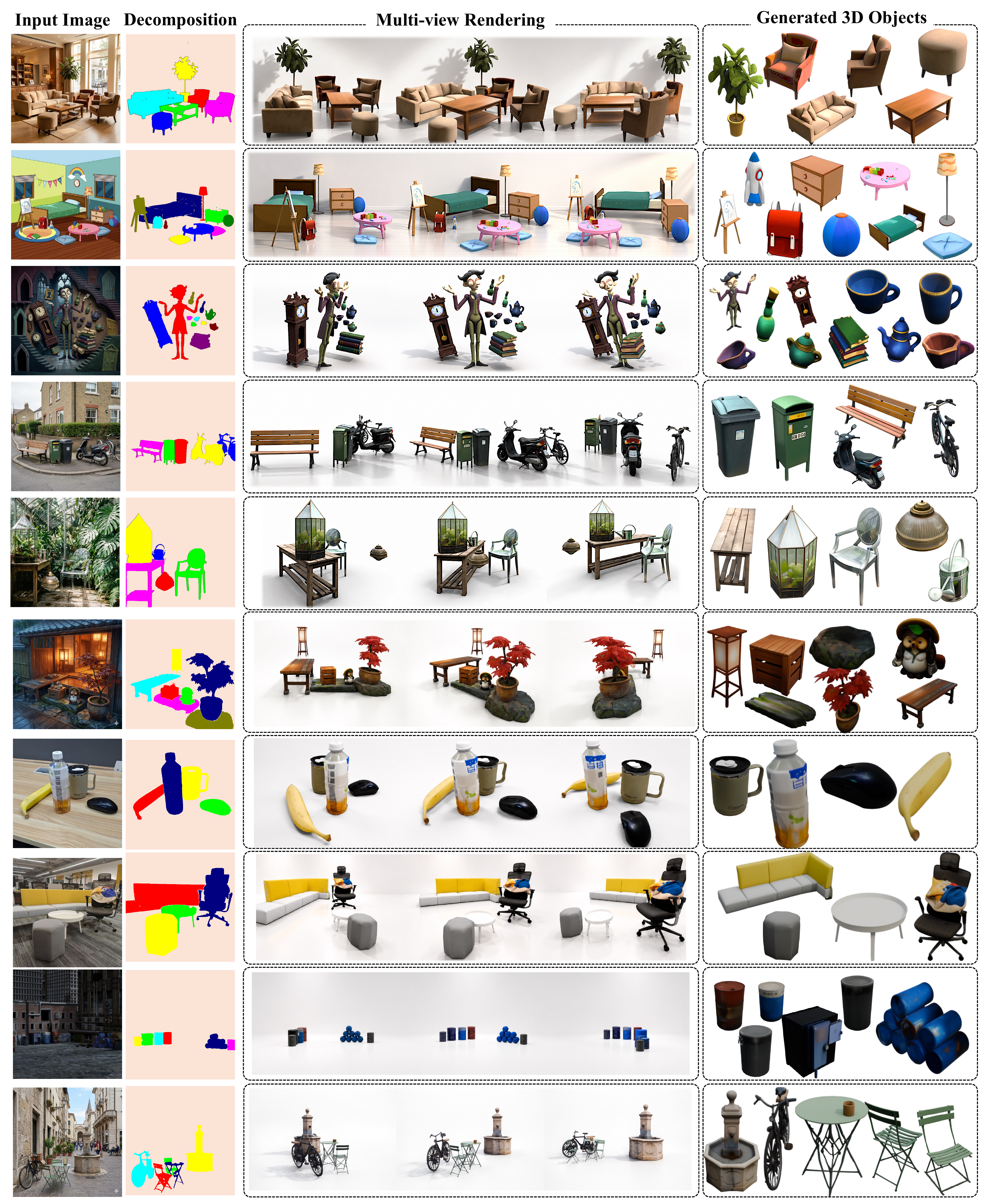}
  \caption{Additional qualitative results on diverse inputs. For each example, we show the input image and the reconstructed compositional 3D scene with object-level assets and novel-view renderings.}
  \label{fig:additional_results}
\end{figure}

\begin{figure}[t]
\centering
  \centering
  \includegraphics[width=0.95\textwidth,height=0.72\textheight,keepaspectratio]{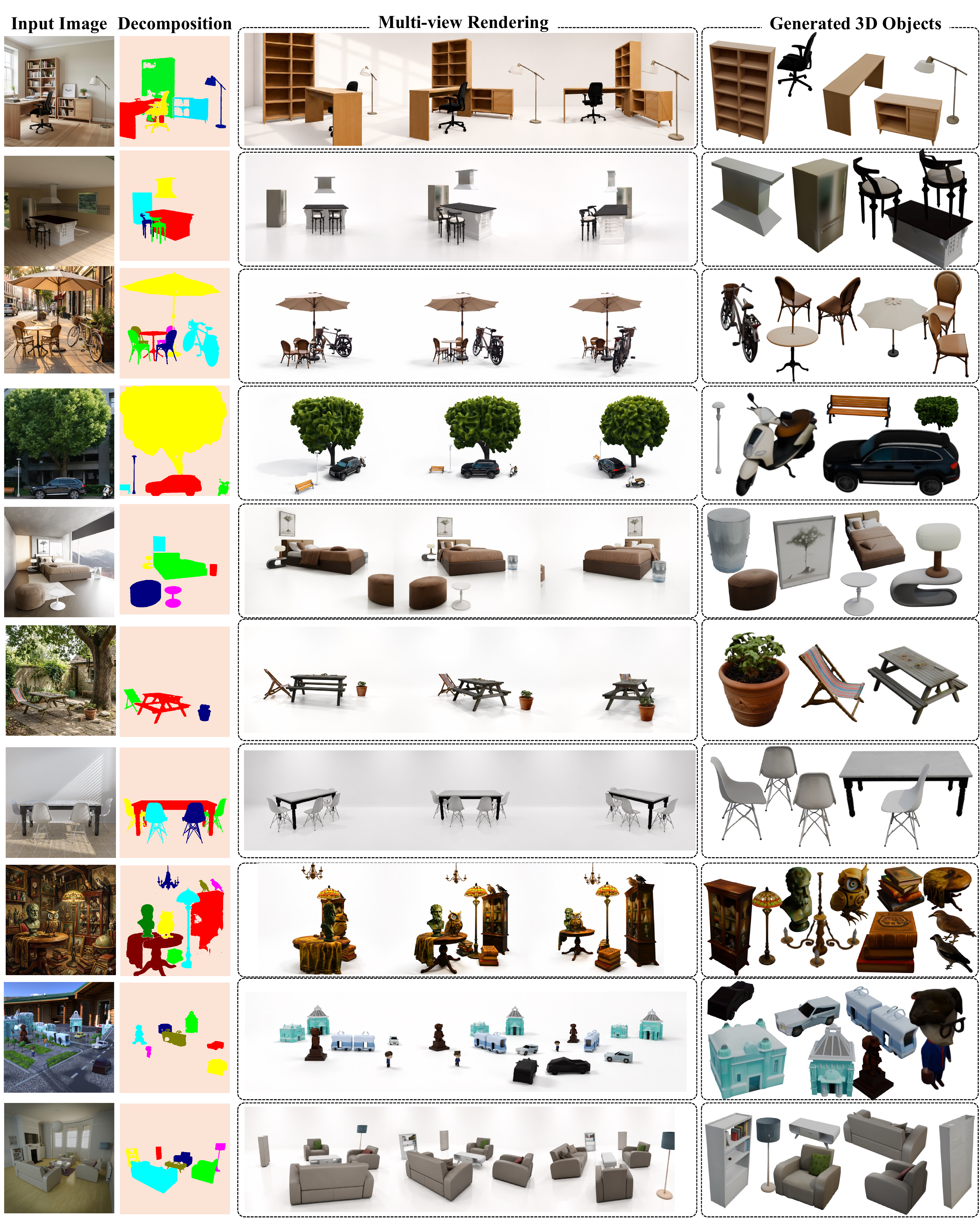}
  \caption{Additional qualitative results, continued. These examples further illustrate that dense CCM--PCM alignment can recover coherent object layouts across indoor, outdoor, synthetic, real, and stylized images.}
  \label{fig:additional_results_2}
\end{figure}

\clearpage
\endgroup

\subsection{\reviewadded{Applications}}
\label{app:applications}
\begin{figure}[htbp]
\centering
\includegraphics[width=0.56\linewidth]{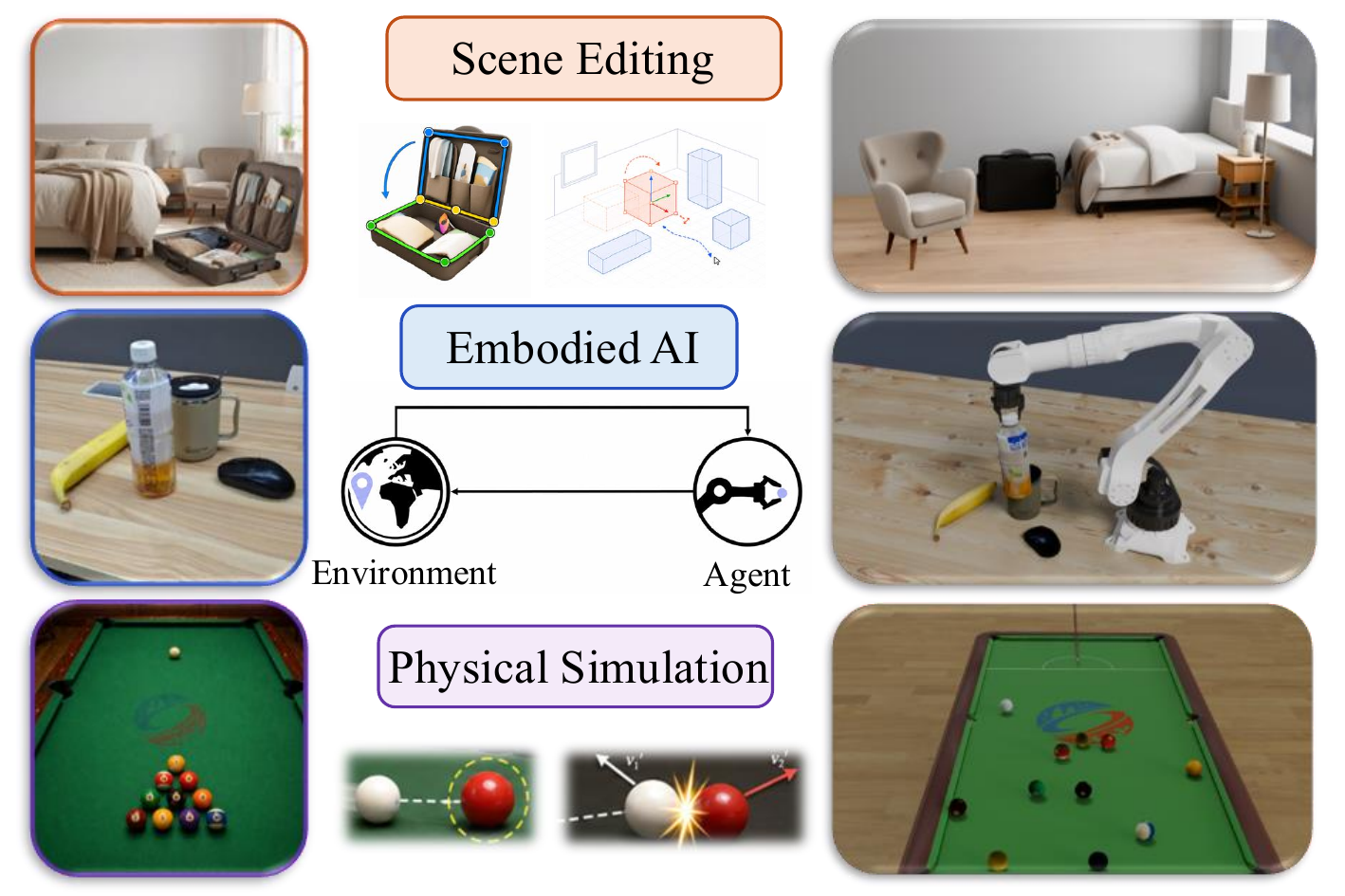}
\caption{\reviewadded{Applications of compositional scene reconstruction: scene editing, embodied AI, and physical simulation.}}
\label{fig:application}
\end{figure}

\reviewadded{Fig.~\ref{fig:application} illustrates scene editing, embodied AI, and physical simulation enabled by our compositional reconstruction. The separate object assets support individual-object editing and export to interactive tools, embodied-AI simulators, and physics engines.}

\FloatBarrier
\begin{reviewmoved}
\section{Related Work}
\label{app:related_work}
\label{app:extended_related_work}

\subsection{Single-Image 3D Generation}



Single-image 3D generation has advanced rapidly at the object level. Early methods often optimize a 3D representation using 2D image priors, such as score distillation~\citep{poole2022dreamfusion, wang2023prolificdreamer,yu2024text} or multi-view consistency constraints~\citep{liu2024syncdreamer,long2024wonder3d}. While these methods can produce plausible object geometry and appearance, they are generally slow and may suffer from multi-view inconsistency. More recent feed-forward and direct 3D generative models learn object-level priors from large-scale 3D data, including large reconstruction models~\citep{hong2023lrm} and diffusion transformers for 3D shapes~\citep{zhang20233dshape2vecset,wu2024direct3d,zhang2024clay,li2025triposg,xiang2025structured, xiang2025native, lai2025hunyuan3d, li2025sparc3d}.
These models greatly improve generation speed and object fidelity.

 Motivated by the strength of object-level 3D priors, recent methods~\citep{huang2025midi,ling2025scene, wang2026scenetransporter, lin2025partcrafter} adapt pre-trained object-centric 3D models to scene-level generation.  In these methods, the entire scene is treated as a single object and normalized into a canonical space. As a result, each individual object occupies only a small fraction of the overall spatial domain. This design introduces an inherent trade-off: while leveraging rich object priors substantially improves the plausibility of generated scenes, each object is represented within only a limited portion of the generation space, which restricts the level of detail and fidelity that can be allocated to individual objects. As scenes become more complex, small, occluded, or thin-structured objects may not receive sufficient effective representation, leading to under-resolved geometry, entangled object boundaries, or inaccurate placement.

Another line of work follows an explicit compositional paradigm, in which each scene is represented as a set of objects together with transformations from the canonical space to the scene space. Based on the source of 3D objects, these methods can be divided into two categories: retrieval-based approaches~\citep{wang2021sceneformer,dai2024automated,tang2024diffuscene} and generation-based approaches. Retrieval-based methods obtain 3D objects from offline libraries, but their generalization to open-set scenarios is limited by asset diversity. Generation- or reconstruction-based methods instead create object assets and differ in the way object geometry and placement are recovered, including direct pose or box prediction~\citep{liu2022towards,detone2026boxer}, joint shape-pose learning~\citep{dahnert2024coherent,meng2025scenegen,chen2026sam}, the use of separate shape and pose modules~\citep{shi2025scenemaker,yao2025cast}, and geometric alignment combined with test-time optimization~\citep{ardelean2025gen3dsr,zhou2024zero,han2025reparo,chen2024comboverse,sautter20253d}. Most of these methods differ in the pipeline used to obtain object shape and pose, while the choice of layout representation remains a largely orthogonal design dimension.

\subsection{3D Layout Generation}

3D layout generation investigates how objects are spatially arranged within a scene. One line of work represents layout using structured or symbolic descriptions, such as scene graphs, language instructions, program-like commands, or relation constraints. Representative methods include 
SceneScript~\citep{avetisyan2024scenescript}, 
LayoutGPT~\citep{feng2023layoutgpt}, 
GALA3D~\citep{zhou2024gala3d}, 
and Holodeck~\citep{yang2024holodeck},  while more recent systems such as LayoutVLM~\citep{sun2025layoutvlm}, Scenethesis~\citep{ling2025scenethesis}, and SceneWeaver~\citep{yang2025sceneweaver} augment this paradigm with foundation-model planning, visual feedback, 
optimization, or physical checking. These representations offer strong semantic controllability and are useful for open-ended scene creation. However, because their layout primitives operate at a coarse semantic level, they are not directly tied to pixel-level geometry, which limits their applicability to image-conditioned scene generation that requires precise geometric correspondence with the input view.


Another common line of work represents layout with sparse object-level scene parameters. 
In indoor scene synthesis, these parameters are often written as box-level object slots, including category, 3D location, size, and yaw orientation~\citep{paschalidou2021atiss,tang2024diffuscene,hu2026mixed,maillard2024debara,yang2024physcene,wu2026caslayout}.
For specific object assets generated in canonical space, an alternative formulation is to express the same layout as an asset-to-scene transformation, typically parameterized by rotation, translation, and optionally scale, as commonly used in object pose estimation~\citep{labbe2022megapose,wen2024foundationpose}. 
This pose-style representation is also widely adopted in compositional scene generation, where individual assets are generated or reconstructed and then placed back into the scene~\citep{wang2021sceneformer,meng2025scenegen,shi2025scenemaker,chen2026sam}.
SceneMaker~\citep{shi2025scenemaker} explicitly adopts this design by decomposing open-set 3D scene generation into dedicated de-occlusion, object generation, and pose estimation modules. SAM3D~\citep{chen2026sam} improves robustness with large-scale visually grounded data, but still represents object layout through sparse pose variables.
Although compact and easy to supervise, such a sparse representation describes only the final placement and provides little intermediate geometric structure for learning. 
In complex scenes with multiple objects, occlusion, perspective ambiguity, and long-tailed spatial configurations, regressing these parameters directly from a single image can be sensitive to outliers and prone to projection drift, object misalignment, or physically implausible layouts.

\reviewadded{Dense coordinate representations provide a different view of layout recovery. While prior work has used such coordinates for category-level pose estimation~\citep{wang2019normalized}, we instead jointly generate each object's geometry and CCM in the same canonical space, integrating correspondence prediction into 3D generation to exploit its learned priors and object-level training data. The generated CCM is aligned with scene-space points to recover object placement through dense correspondences. Compared with sparse box or pose variables, this pixel-aligned and bounded coordinate field enables layout learning from object-level assets without requiring scene-level layout annotations for those assets. At inference time, it yields an over-determined alignment problem that is more robust to local prediction errors than regressing a few global pose parameters.}


\subsection{Multi-Modal Generative Models}
\label{sec:related_multimodal}

Diffusion Transformers (DiTs)~\citep{peebles2023scalable} have become a strong backbone for generative modeling due to their scalability and simple transformer design. Multimodal variants further study how heterogeneous token types can interact within one generative model. MMDiT~\citep{esser2024scaling} extends DiTs to text-image generation with modality-specific weights and joint attention. Mixture-of-Transformers (MoT)~\citep{liang2024mixture} demonstrates that decoupling parameters by modality while sharing self-attention enables efficient multi-modal learning. BAGEL~\citep{deng2025emerging} adopts this design for unified understanding and generation. 

Recent multimodal diffusion models further extend this idea from text-image generation to the joint generation of multiple output modalities, including text, image, audio, video, and action~\citep{li2025omniflow,won2025dual,chen2026skyreels,liu2025javisdit,hacohen2026ltx,guo2026alive,wang2026apollo,zhao2025uniform,qiang2026mm,wang2024av}.
For example, OmniFlow~\citep{li2025omniflow} employs modality-specific processing with shared attention for unified text, image, and audio generation, while DUST~\citep{won2025dual} uses separate action and visual streams with shared attention for joint action-video prediction.
Together, these works show that modality-aware processing with shared attention is effective for multimodal co-generation.

In our setting, we treat canonical geometry and dense layout as two output modalities. We jointly generate canonical 3D object geometry and dense 2D layout maps. These two outputs have different spatial structures: geometry is represented in canonical 3D space, while the CCM is represented as a pixel-aligned 2D map. Simply concatenating their tokens may ignore this structural difference, while fully separate models would miss the strong dependency between object shape and layout. We therefore introduce a Mixture-of-Transformers (MoT) architecture for geometry-layout co-generation, keeping geometry and layout as separate expert streams while allowing them to exchange information through shared self-attention and a shared geometry-layout positional space. \reviewadded{To the best of our knowledge, we are the first to jointly denoise object geometry in a 3D latent space and pixel-aligned canonical coordinate maps in 2D image space within a Mixture-of-Transformers architecture.}

\end{reviewmoved}

\FloatBarrier
\section{\reviewadded{Method Details}}
\label{app:method_details}
\subsection{\reviewadded{Architecture and Image Conditioning}}
\label{app:architecture_details}
\begin{reviewmoved}
Both the geometry and layout experts follow the TRELLIS~\citep{xiang2025structured} transformer design, using 24 layers, hidden dimension 1024, and 16 attention heads. The geometry stream operates on $8\times8\times8$ sparse shape latents with 8 channels, encoded by a sparse-structure VAE with latent dimension 8. The layout stream predicts CCMs at $296\times296$ crop-space resolution and tokenizes them with a strided convolution of patch size 8.

\begin{wrapfigure}{r}{0.50\textwidth}
\setlength{\abovecaptionskip}{5pt}
\reviewmovedcolor
\centering
  \includegraphics[width=\linewidth]{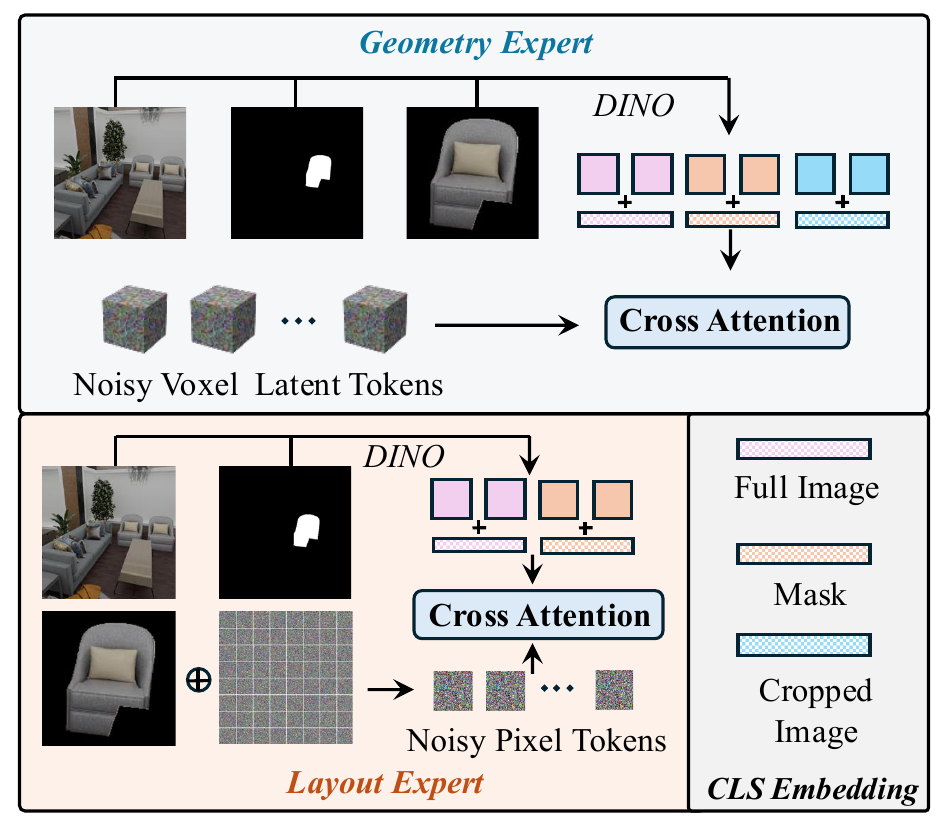}
  \caption{Image condition injection for the geometry and layout experts. The geometry expert uses full-image, mask, and object-crop features through cross-attention, while the layout expert  concatenates the cropped RGB image with the noisy CCM for pixel-aligned local conditioning.}
  \label{fig:image_cond}
\end{wrapfigure}
\leavevmode\textbf{\reviewadded{Geometry implementation details.}} The timestep information is injected through adaptive layer normalization and a gating mechanism~\citep{peebles2023scalable}. After denoising, the latent grid is decoded back to the canonical occupancy grid $S_k$.


\textbf{\reviewadded{Geometry image conditioning.}} Image conditions are injected through cross-attention layers as keys and values, following~\citep{xiang2025structured}. As shown in the upper part of Fig.~\ref{fig:image_cond}, we condition the geometry expert on DINOv2~\citep{oquab2023dinov2} features extracted from the full image, the object mask, and the cropped object image. Learned type embeddings are added to distinguish these feature sources.


\textbf{\reviewadded{Layout image conditioning.}} Unlike geometry, the CCM is fully pixel-aligned with the cropped image. We therefore concatenate the cropped RGB image with the noisy CCM along the channel dimension as a local image condition. To further inject global context, we encode the full image and mask with DINOv2, add learned type embeddings, and incorporate the resulting features into the DiT through cross-attention, as illustrated in Fig.~\ref{fig:image_cond}.
\end{reviewmoved}

\subsection{\reviewadded{Robust Geometric Alignment}}\label{app:alignment}
\label{sec:supp_alignment}

This section details the post-processing step used to recover object transformations from predicted CCMs and estimated PCMs. The goal is to estimate a similarity transformation $(s,R,t)$ that maps canonical object coordinates to the camera-space scene frame:
\begin{equation}
    p^{\mathrm{cam}} = sRp^{\mathrm{can}} + t,
\end{equation}
where $s\in\mathbb{R}^{+}$ is an isotropic scale, $R\in SO(3)$ is a rotation matrix, and $t\in\mathbb{R}^{3}$ is a translation vector. In our implementation, the canonical object frame is Z-up with coordinates normalized to $[-0.5,0.5]$, and the camera frame follows the OpenGL convention: $+X$ points right, $+Y$ points up, and $-Z$ points forward.

\paragraph{Camera-space point map.}
\reviewadded{We estimate a depth map $D$ from the input image using Pixel-Perfect Depth~\citep{xu2025pixel}. Given the estimated horizontal field-of-view $\theta_{\mathrm{fov}}$, we unproject each pixel $(u,v)$ with depth $d$ into camera space:}
\begin{equation}
    X = \frac{(u-c_x)d}{f_x},\quad
    Y = -\frac{(v-c_y)d}{f_x},\quad
    Z = -d,
\end{equation}
where $f_x=\frac{W}{2}\cot(\frac{\theta_{\mathrm{fov}}}{2})$ and $(c_x,c_y)=(\frac{W}{2},\frac{H}{2})$. This yields a camera-space PCM $P\in\mathbb{R}^{H\times W\times 3}$.


\paragraph{Valid correspondence extraction.}
For each object $k$, we first paste the predicted crop-space CCM back to the full image using the inverse crop transform. We then construct a valid correspondence set
\begin{equation}
    \Omega_k =
    \{u \mid M_k(u)=1,\ C_k(u)\ \text{is valid},\ P(u)\ \text{is valid}\},
\end{equation}
where $M_k$ is the object mask. A pixel is retained only when it is inside the instance mask, marked valid by the monocular geometry estimator, and is not a near-origin low-confidence CCM prediction. We use $\|C_k(u)\|_2 > \tau_{\mathrm{origin}}$ with $\tau_{\mathrm{origin}}=0.02$ for the near-origin check.

\paragraph{Depth-edge filtering.}
Pixels near depth discontinuities are often unreliable because they lie on occlusion boundaries or mask boundaries. We detect such pixels using a relative depth range in a local $(2r+1)^2$ neighborhood:
\begin{equation}
    \rho_u =
    \frac{d_{\max}^{(r)} - d_{\min}^{(r)}}{|d_{\mathrm{mean}}^{(r)}|+\epsilon},
\end{equation}
where $r=3$ by default. Pixels with $\rho_u>\tau_{\mathrm{edge}}$ are treated as depth-edge pixels, with $\tau_{\mathrm{edge}}=0.05$ by default. We then dilate the edge mask to remove adjacent unreliable pixels. Since monocular depth may be non-metric, this relative criterion makes filtering scale-invariant. As a safeguard, if depth-edge filtering removes more than $70\%$ of initially valid pixels, we relax or disable the edge filter to avoid discarding too many correspondences.

\paragraph{RANSAC hypothesis generation.}
Both CCM prediction and monocular geometry estimation can contain local errors. Direct least-squares alignment over all pixels may therefore be biased by noisy depth, mask boundary errors, or incorrect correspondences. We use RANSAC~\citep{fischler1981random} to robustly estimate the transformation. If the number of valid correspondences exceeds $N_{\max}=3000$, we uniformly subsample $N_{\max}$ correspondences to keep per-iteration residual computation bounded.

Rather than using a fixed inlier threshold, we set the threshold adaptively according to the spatial extent of the camera-space points:
\begin{equation}
    \delta =
    \alpha
    \left\|
    \max(p^{\mathrm{cam}}) - \min(p^{\mathrm{cam}})
    \right\|_2,
\end{equation}
where $\alpha=0.02$ by default. This makes the inlier threshold scale-invariant with respect to the non-metric scale of monocular geometry.

For each RANSAC iteration, we randomly sample $m=6$ correspondences and estimate a candidate similarity transform using the closed-form Umeyama algorithm~\citep{umeyama1991least}. Given a candidate $(s,R,t)$, the residual of correspondence $i$ is
\begin{equation}
    r_i = \left\| P_i - (sRC_i+t) \right\|_2 .
\end{equation}
The inlier set is defined as
\begin{equation}
    \mathcal{I} = \{i\in\Omega_k \mid r_i < \delta \}.
\end{equation}
We use adaptive early stopping with the standard RANSAC criterion. Given the current best inlier rate $\hat{w}=n^*/N$, where $n^*$ is the best inlier count found so far and $N=|\Omega_k|$  is the number of sampled correspondences, the number of iterations needed for $99\%$ success probability is estimated as
\begin{equation}
    \left\lceil
    \frac{\log(0.01)}{\log(1-\hat{w}^{m})}
    \right\rceil .
\end{equation}
The RANSAC loop terminates once this bound is reached.

\paragraph{Final refinement.}
After RANSAC, we select the hypothesis with the best inlier support and iteratively refine the transformation. At each refinement step, we refit the similarity transformation using the current inlier set and recompute inliers using the adaptive threshold $\delta$. We run up to $L=5$ refinement rounds and stop early if the inlier count no longer increases. The final transform is obtained by solving
\begin{equation}
    s^{*},R^{*},t^{*}
    =
    \arg\min_{s,R,t}
    \sum_{i\in\mathcal{I}}
    \left\|
    P_i - (sRC_i+t)
    \right\|_2^2 .
\end{equation}
The resulting transformation is applied to all generated canonical object geometry.

\begin{table}[htb]
\reviewtableclean
\setlength{\belowcaptionskip}{6pt}
\caption{Robust alignment hyperparameters used in our implementation.}
\label{tab:supp_alignment_hparams}
\centering
\begin{tabular}{lc}
\toprule
\textbf{Parameter} & \textbf{Value} \\
\midrule
Canonical coordinate range & $[-0.5,0.5]$ \\
Near-origin CCM threshold $\tau_{\mathrm{origin}}$ & 0.02 \\
Depth-edge radius $r$ & 3 \\
Depth-edge threshold $\tau_{\mathrm{edge}}$ & 0.05 \\
Maximum correspondences per object $N_{\max}$ & 3000 \\
RANSAC sample size $m$ & 6 \\
Adaptive threshold scale $\alpha$ & 0.02 \\
RANSAC success probability & 99\% \\
Refinement rounds $L$ & 5 \\
Minimum inliers $n_{\min}$ & 10 \\
\bottomrule
\end{tabular}
\end{table}

\paragraph{Joint shared-up variant.}
For multi-object scenes, our implementation also supports a joint solving mode that constrains all objects to share the same up direction in camera space. Under this constraint, each object rotation is decomposed into a shared up direction and an object-specific in-plane angle. Scale and translation are computed in closed form during optimization. This variant can improve physical plausibility for scenes where objects rest on a common ground plane, although the main experiments use the per-object solver unless otherwise specified.

\subsection{\reviewadded{Rectified Flow Background}}
\label{app:network_details}
\begin{reviewmoved}
\textbf{Rectified Flow Models.} Rectified flow models use a linear interpolation forward process, where a timestep $t$ is used to interpolate between a data sample and Gaussian noise $\bm{\epsilon}$, formulated as
\begin{equation}
    \bm{x}(t) = (1-t) \bm{x}_0 + t\bm{\epsilon}.
\end{equation}
The reverse process is modeled as a time-dependent vector field, $\bm{v}(\bm{x},t) = \nabla_t \bm{x}$ that transports noisy samples back to the data distribution. This can be learned by minimizing the conditional flow matching objective~\citep{lipman2022flow}:
\begin{equation}
    \mathcal{L}_{CFM}^{t,\bm{\epsilon}}(\bm{x}) = \mathbb{E}_{t,\bm{x}_0,\bm{\epsilon}}|| \bm{v}(\bm{x},t) - (\bm{\epsilon} - \bm{x}_0)||_2^2.
\end{equation}
\end{reviewmoved}

\FloatBarrier
\section{\reviewadded{Data and Implementation Details}}
\label{app:implementation}
\subsection{\reviewadded{Data Construction Details}}\label{app:data_construction}

\label{sec:supp_data}

Our training data construction follows the two-stage strategy described in the main paper.

\paragraph{Object-level pre-training data.}
We select 60K object assets from Objaverse~\citep{deitke2023objaverse} and render 1M object-centric views. These renderings provide direct supervision for canonical geometry and CCMs, since the visible surface coordinate at each pixel can be obtained from the rendered canonical object. This object-level supervision is sufficient for learning dense canonical correspondence before the model observes complex scene layouts.

\paragraph{Background-completed object views.}
To reduce the domain gap between isolated object renderings and real scene images, we sample a subset of the rendered views for background completion. We use FLUX2.0~\citep{flux-2-2025} to synthesize plausible backgrounds while preserving the foreground object, followed by a foreground consistency check to reject examples where the object content is changed. Because background completion is computationally expensive, we do not process all 1M rendered views. The final background-completed subset contains 20K photo-realistic object views.

These background-completed object views are also used when training the Raw and Coord Cube layout baselines. Unlike CCM, these baselines predict scene-space quantities and therefore require scene-like supervision. For each background-completed object view, we estimate a point cloud using MoGe~\citep{wang2025moge} and align the canonical object to this estimated point cloud.
The resulting object-to-point-cloud transformation provides the \reviewadded{estimated} scene-space pose or scene-space coordinate target used to supervise the Raw and Coord Cube representations.

\paragraph{Scene-level fine-tuning data.}
We fine-tune on 20K 3D-FRONT~\citep{fu20213d} scene views. The purpose of this stage is not to replace object-level pre-training, but to adapt the model to real scene phenomena such as occlusion, partial visibility, perspective variation, and amodal object completion. During training, we sample object instances from scene views and use their masks to construct object crops and corresponding CCM supervision.

\subsection{\reviewadded{Training Settings}}
\label{app:training_settings}
\begin{reviewmoved}
We train the generative model with rectified flow matching~\citep{lipman2022flow} and \texttt{logit\_normal} timestep sampling. For the geometry expert, we apply 10\% condition dropout during training and use classifier-free guidance~\citep{ho2022classifier} at inference. For the layout expert, we do not use classifier-free guidance, as we find deterministic image conditioning more stable for dense coordinate prediction.

We initialize the geometry stream from the TRELLIS-image-large~\citep{xiang2025structured} object generation model, which provides a strong object-level geometry prior.
Pre-training is conducted for 100K steps on 32 NVIDIA A100-80G GPUs with batch size 4 per GPU.
We use AdamW~\citep{loshchilov2017decoupled} with $\beta_1=0.9$ and $\beta_2=0.999$. The base learning rate is set to $1 \times 10^{-4}$. Fine-tuning is performed on 8 NVIDIA A100-80G GPUs with batch size 4.
The learning rate is set to $1 \times 10^{-4}$ and linearly decayed to $5 \times 10^{-5}$.
We use BFloat16 mixed precision and clip gradients with a maximum norm of 1.0.
\end{reviewmoved}

\subsection{\reviewadded{Input Preprocessing and Inference}}
\label{app:inference_details}
\noindent\reviewadded{All input images and object masks are resized to $518\times518$. At test time, given a single RGB image and ground-truth instance masks, we run 30-step Euler sampling with classifier-free guidance scale 3.0 to jointly predict per-object geometry and CCMs. We estimate camera-space point maps using Pixel-Perfect Depth~\citep{xu2025pixel}. We recover object placement using the scene assembly procedure described in Sec.~\ref{sec:scene_assemble}; the robust alignment details are provided in Appendix~\ref{app:alignment}. Our geometry expert predicts only the sparse voxel structure. To obtain final meshes for rendering and evaluation, we reuse the second-stage mesh generation/refinement module from SAM3D~\citep{chen2026sam}. Our contribution focuses on geometry-layout co-generation and correspondence-based layout recovery rather than mesh refinement.}

\begin{reviewaddedblock}
\subsection{Automatic VLM--SAM3 Instance Segmentation}
\label{app:auto_segmentation}

We use a VLM-guided SAM3~\citep{carion2026sam} pipeline adapted from REST3D~\citep{ma2026rest3d} to obtain visible instance masks when they are not supplied. We use GPT-5.6 as the VLM throughout this pipeline. Figure~\ref{fig:agentic_segmentation} illustrates the pipeline and how candidate recycling recovers missed instances.

\begin{figure}[htbp]
\centering
\includegraphics[width=\linewidth]{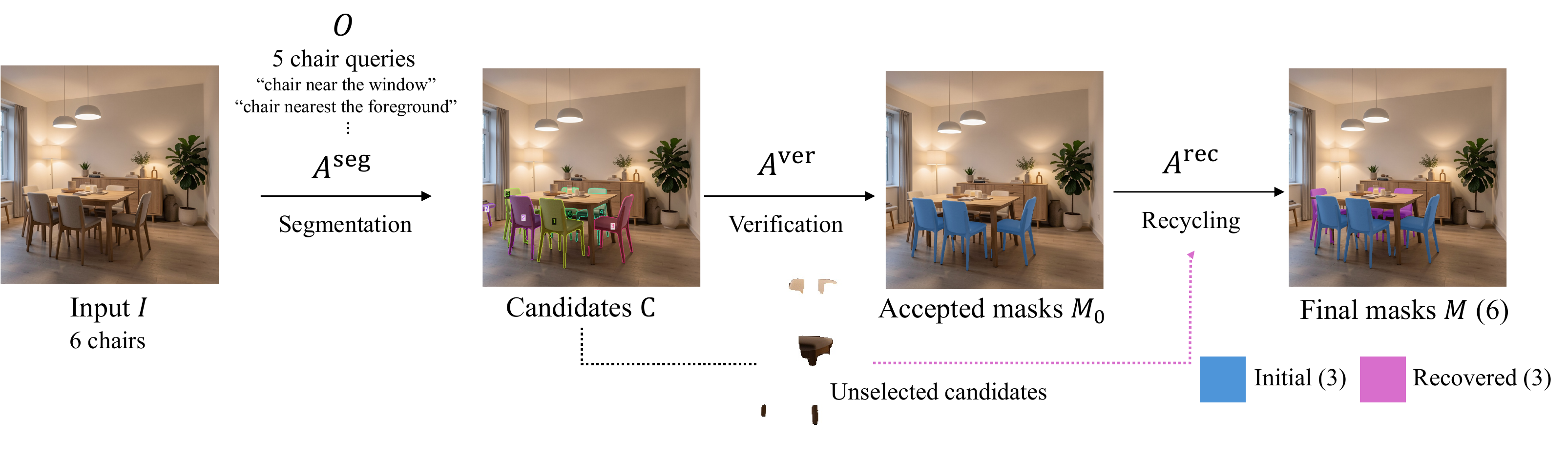}
\caption{\reviewadded{Automatic instance segmentation with candidate recycling. The scene contains six chairs, but the object list includes only five chair descriptions. The candidate panel shows one representative SAM3 response. Selection, membership verification, and cross-query deduplication yield three initial chair masks (blue). Recycling recovers three additional chairs (purple) from cached candidates, producing six final chair masks. The dashed path illustrates an unselected candidate reused during recycling.}}
\label{fig:agentic_segmentation}
\end{figure}

\paragraph{Object queries and segmentation.}
Given an image $I$, a VLM produces an object list $\mathcal{O}$ with appearance and spatial descriptions. Room-scale inputs focus on major foreground objects, while tabletop inputs include the supporting surface and individually identifiable objects on it. The segmentation agent $A^{\mathrm{seg}}$ queries SAM3 using these descriptions. Within this agent, a VLM examines candidate overlays to select matching masks or revise the query. Candidates generated across queries and retries are retained in a shared pool $\mathcal{C}$.

\paragraph{Instance membership verification.}
The verifier $A^{\mathrm{ver}}$ distinguishes between single-object and group queries. For a single-object query with multiple selected masks, it checks whether they depict visible parts of the same physical instance before merging them. This prevents, for example, the backrest of a neighboring chair from being merged into the target chair. For a group query, each selected mask is checked for membership in the requested group and correspondence to an independent instance; valid instances are retained separately. Cross-query deduplication then yields the initial accepted masks $\mathcal{M}_0$.

\paragraph{Unselected-candidate recycling.}
The initial object list may omit instances, and spatial or appearance descriptions do not guarantee a one-to-one mapping to physical objects. Different queries can select the same subset of similar objects while leaving valid candidates unselected. Removing duplicate selections alone does not recover these missing instances.

The recycling agent $A^{\mathrm{rec}}$ revisits $\mathcal{C}$ using the accepted masks $\mathcal{M}_0$. It first verifies whether a candidate can replace a partial mask of the same instance. After updating the accepted masks, overlap filtering removes duplicate and already-covered candidates. VLM identity verification then examines the remaining candidates using scene context and mask-only crops, rejecting background regions, unrelated fragments, and masks spanning multiple objects. Verified new instances are added, and final deduplication produces $\mathcal{M}$. In Fig.~\ref{fig:agentic_segmentation}, the five chair descriptions yield three distinct accepted chairs, while recycling recovers the other three from cached candidates. This step requires no additional SAM3 calls.

Detailed prompts and configuration settings will be released with the code. The quantitative comparisons reported in this paper use ground-truth instance masks.
\end{reviewaddedblock}

\subsection{\reviewadded{Details of Layout Representations}}
\label{sec:supp_layout_rep}

The main paper compares three layout representations: \emph{Raw}, \emph{Coord Cube}, and \emph{CCM with PCM}. All three variants use the same geometry branch and the same overall geometry-layout co-generation architecture unless otherwise specified. They differ only in the target predicted by the layout branch and in the way the final object-to-scene transformation is recovered.

Table~\ref{tab:supp_layout_tokenization} summarizes the tokenization used by the layout branch for the three representations. The raw pose representation is allocated 8 layout tokens. Coord Cube uses a $32\times32\times32$ canonical grid and a $2\times2\times2$ patch size, producing $16\times16\times16=4096$ layout tokens. CCM is predicted at $296\times296$ resolution in crop space and tokenized with an $8\times8$ patch size, producing $37\times37=1369$ layout tokens.

\begin{table}[htb]
\reviewtableclean
\setlength{\belowcaptionskip}{6pt}
\caption{Layout representation tokenization used in the ablation study.}
\label{tab:supp_layout_tokenization}
\centering
\begin{tabular}{lccc}
\toprule
\textbf{Representation} & \textbf{Resolution} & \textbf{Patch size} & \textbf{Tokens} \\
\midrule
Raw pose & -- & -- & 8 \\
Coord Cube & $32\times32\times32$ & $2\times2\times2$ & 4096 \\
CCM with PCM & $296\times296$ & $8\times8$ & 1369 \\
\bottomrule
\end{tabular}
\end{table}

\paragraph{Raw pose representation.}
The raw baseline directly predicts an object-to-scene transformation using 8 layout tokens:
\begin{equation}
    \reviewadded{T=(s,R(q),t),}
\end{equation}
where $s \in \mathbb{R}^{+}$ is the object scale, \reviewadded{$q \in \mathbb{R}^{4}$ is the rotation quaternion with corresponding rotation matrix $R(q)\in SO(3)$}, and $t \in \mathbb{R}^{3}$ is the object translation in the scene frame. This representation is compact, but it is also sparse and unbounded. Since the predicted variables directly live in the scene coordinate frame, training requires scene-level pose supervision.

\paragraph{Coord Cube representation.}
Coord Cube densifies scene-space prediction. We first construct a fixed canonical coordinate tensor $X\in\mathbb{R}^{32\times32\times32\times 3}$ by uniformly sampling a regular grid in the canonical object space. The tensor is patchified with a $2\times2\times2$ patch size, yielding 4096 layout tokens. Flattening $X$ gives $J=32^3$ canonical points $\{x_j\}_{j=1}^{J}$. Instead of directly predicting $(s,R,t)$, the layout branch predicts their corresponding scene-space locations $\{\hat{y}_j\}_{j=1}^{J}$, equivalently a scene-space coordinate tensor $\hat{Y}\in\mathbb{R}^{32\times32\times32\times 3}$. The object transformation is then recovered by solving
\begin{equation}
    s^{*},R^{*},t^{*}
    =
    \arg\min_{s,R,t}
    \sum_{j=1}^{J}
    \left\|
    \hat{y}_j - (sRx_j+t)
    \right\|_2^2 .
\end{equation}
Compared with raw pose regression, Coord Cube provides a denser prediction target and is more robust to small local errors. However, the predicted coordinates $\hat{y}_j$ are still scene-space quantities. Their range depends on camera pose, object depth, scene scale, and object location. Therefore, Coord Cube remains unbounded and still requires scene-level supervision to learn accurate scene-space coordinates.

\paragraph{CCM with PCM.}
Our representation inverts the prediction direction. Instead of predicting where canonical points should appear in the scene, the layout branch predicts which canonical point is observed by each visible image pixel. For an object $k$, the Canonical Coordinate Map (CCM) $C_k$ stores a canonical coordinate $C_k(u)$ at each visible pixel $u$. We predict CCM at $296\times296$ crop-space resolution and use an $8\times8$ patch size, resulting in 1369 layout tokens. After the predicted crop-space CCM is mapped back to the full image, each valid pixel provides a dense correspondence between canonical object space and the scene frame:
\begin{equation}
    C_k(u) \leftrightarrow P(u),
\end{equation}
where $P(u)$ is the scene-space point from the Point Cloud Map (PCM). The object transformation is recovered by robust geometric alignment over these dense correspondences.

\paragraph{Training of layout baselines.}
For a fair comparison, the Raw and Coord Cube variants share the same network architecture as the CCM variant. Since Raw and Coord Cube require scene-space supervision, we train them using the background-completed object views and scene-level data described in Sec.~\ref{sec:supp_data}. 
In addition, our CCM-based layout recovery explicitly uses a monocular geometry prior through PCM alignment. To ensure that the comparison focuses on the layout representation rather than access to monocular geometry, we replace the DINOv2 features in the layout branch of the Raw and Coord Cube variants with monocular geometry estimation features~\citep{wang2025moge}, following the design choice used in SAM3D~\citep{chen2026sam}. This gives the scene-space baselines access to comparable depth and geometry cues and avoids underestimating their performance due to weaker image conditioning.

\FloatBarrier
\section{\reviewadded{Evaluation Protocol}}
\label{app:evaluation_details}

\label{app:evaluation_protocol}

\subsection{\reviewadded{Evaluation Datasets}}
\label{app:evaluation_datasets}
\noindent\reviewadded{Training data curation is described in Sec.~\ref{sec:training_pipeline}. 3D-Future Scene~\citep{fu2021future} is used as training data by prior methods such as SceneGen~\citep{meng2025scenegen}. This benchmark is distinct from the 3D-FRONT views rendered by us for training, and the two data sources have different rendering and annotation distributions. Since the masks in 3D-Future Scene have varying quality and often contain noise, we select 50 examples to evaluate our ability to handle indoor scenes under this established setting. Our second benchmark is constructed from BlendSwap~\citep{blendswap}, an online repository for Blender assets. Compared with 3D-Future Scene, this benchmark provides higher-quality annotations and more diverse scene styles, and we therefore use it as our primary benchmark for evaluating reconstruction quality and cross-domain generalization. For qualitative evaluation, in-the-wild inputs include real, photorealistic, and stylized images collected from the web or generated by Gemini 3 Pro Image (Nano Banana Pro) and FLUX2~\citep{flux-2-2025}.}

\subsection{\reviewadded{Metric Definitions and Computation}}
\label{app:metric_definitions}
\label{app:metric_computation}

\begin{reviewmoved}
\leavevmode\reviewadded{\textbf{Object geometry metrics.}}\space
For object geometry, we uniformly sample point clouds from the generated and ground-truth asset surfaces, normalize each matched object pair, and align them using a robust ICP procedure~\citep{ling2025scene}. \reviewadded{We report CD, FS@0.1, and EMD. Earth Mover's Distance (EMD) complements nearest-neighbor metrics by measuring the transport cost between point distributions.}
\end{reviewmoved}

\begin{reviewmoved}
\leavevmode\reviewadded{\textbf{Scene layout metrics.}}\space
For layout metrics, we normalize the predicted and ground-truth scenes by their scene-level bounding boxes, merge all object point clouds in each scene, and estimate a single robust ICP transformation from the predicted scene to the ground truth. This transformation is applied uniformly to all predicted objects. 3D-IoU is then computed between matched object axis-aligned bounding boxes in the aligned scene space.

ADD-S~\citep{xiang2017posecnn} is computed on posed predicted and ground-truth object point sets. Following SAM3D~\citep{chen2026sam}, we use the symmetric formulation:
\[
\mathrm{ADD}(\mathcal{A}, \mathcal{B})
=
\frac{1}{|\mathcal{A}|}
\sum_{\mathbf{x}\in\mathcal{A}}
\min_{\mathbf{y}\in\mathcal{B}}
\|\mathbf{x}-\mathbf{y}\|_2,
\]
\[
\mathrm{ADD\text{-}S}
=
\frac{
\mathrm{ADD}(\mathcal{M}, \mathcal{M}_{\mathrm{gt}})
+
\mathrm{ADD}(\mathcal{M}_{\mathrm{gt}}, \mathcal{M})
}{2d},
\]
where $\mathcal{M}$ and $\mathcal{M}_{\mathrm{gt}}$ are the predicted and ground-truth posed point clouds, and $d$ is the diameter of $\mathcal{M}_{\mathrm{gt}}$. For 2D-IoU, we render predicted and ground-truth object silhouettes from the calibrated input camera and compute their mask overlap. Finally, ICP-Rot is computed per object by centering the predicted and ground-truth point clouds, scaling them by the ground-truth object diameter, and reporting the residual rotation angle estimated by point-to-point ICP in this normalized object space.
\end{reviewmoved}

\subsection{\reviewadded{Normalization and Alignment}}
\label{app:evaluation_alignment}
Before evaluation, predictions from all methods are converted into a common evaluation coordinate frame using the corresponding camera calibration and up-axis convention. This ensures that subsequent point-cloud registration compares geometry under consistent camera and axis conventions, rather than compensating for coordinate-system differences during alignment.

\noindent\textbf{Robust ICP implementation.}
We use robust ICP as the common alignment primitive for object geometry evaluation and scene layout evaluation. Our implementation follows the high-level strategy of I-Scene~\citep{ling2025scene}, with additional initialization steps to reduce sensitivity to local minima. We first perform a yaw-sweep initialization around the evaluation up axis. Candidate yaw rotations are pre-scored using a trimmed symmetric Chamfer distance on downsampled point clouds, and the best candidates are refined with short point-to-point ICP. Starting from the best initialization, we run coarse-to-fine registration: a coarse point-to-point ICP stage on voxel-downsampled points, followed by a fine full-resolution refinement using point-to-plane ICP with a robust loss when normals are available.

\subsection{\reviewadded{3D--2D Correspondence Evaluation}}
\label{app:correspondence_protocol}
\reviewadded{For the comparison in Sec.~\ref{sec:correspondence_analysis}, Tab.~\ref{tab:3d-2d-correspondence} reports four settings: CUPID-(Mesh+GT), Ours-(Mesh+GT), Ours-(CCM+GT), and Ours-(CCM+Mesh).}
\begin{reviewmoved}
\emph{Mesh+GT} follows the CUPID evaluation setting: we render the final generated and posed mesh, back-project its rendered depth into 3D points, and compute the loss against 3D points back-projected from the ground-truth rendered depth. \emph{CCM+GT} directly compares the 3D points induced by our predicted CCM with the ground-truth depth back-projected points. \emph{CCM+Mesh} compares the CCM-induced 3D points with the depth back-projected points from our final generated and posed mesh, measuring the internal consistency between our layout and geometry branches.
\end{reviewmoved}

\FloatBarrier
\section{\reviewadded{Limitations and Future Work}}
\label{app:limitations}
\begin{reviewmoved}
\begin{wrapfigure}{r}{0.50\textwidth}
\setlength{\abovecaptionskip}{5pt}
  \reviewmovedcolor
  \includegraphics[width=\linewidth]{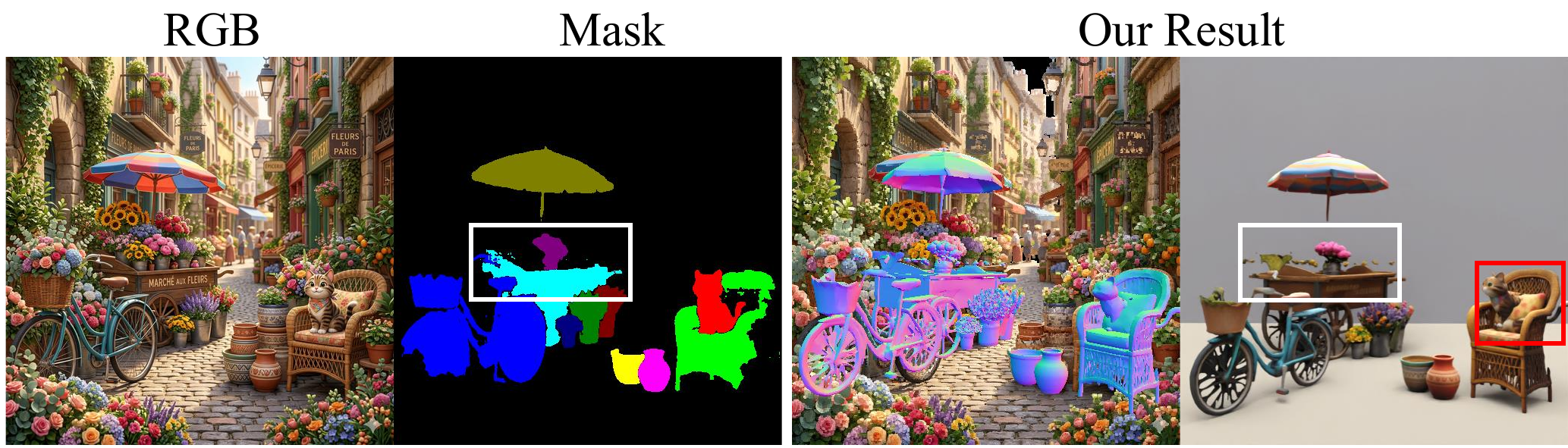}
  \caption{Representative failure cases. Independent per-object reconstruction may cause object intersections, and noisy masks can introduce floaters or missing object parts.}
  \label{fig:failure}
\end{wrapfigure}

Although our method achieves high-quality scene generation results on most images, we observed several limitations during testing. One major issue is that the current pipeline processes each object independently. Therefore, when objects are close to each other or when the generated object geometry is not sufficiently accurate, self-intersections may appear in the final reconstructed scene. For example, as highlighted by the red box in Fig.~\ref{fig:failure}, the cat's tail intersects with the chair. This issue could potentially be alleviated by joint multi-object generation~\citep{huang2025midi} or by introducing additional post-processing optimization~\citep{yao2025cast}.

In addition, our method requires both an RGB image and the corresponding object masks as input. However, obtaining accurate masks from real images is non-trivial in practice. As shown in the white box of Fig.~\ref{fig:failure}, for complex scenes, the masks often contain noise, which can lead to floaters in the final results. Reducing the dependence on masks and enabling the model to recover objects directly from image semantics remains an important direction for future work.

Another characteristic of Mira-Scene is that it mainly learns 2D-3D correspondence, while relying on monocular geometry estimation methods to obtain global scene geometry. Although we find that current geometry estimation methods~\citep{wang2025moge, xu2025pixel} are sufficiently robust and can produce reasonable results in most cases, they may still generate thin, sheet-like point clouds for some 2D cartoon-style images, which consequently affects the quality of our results. This limitation is expected to be mitigated as stronger geometry estimation methods become available.

Finally, the amount of training data used in our current model is substantially smaller than that used by SAM3D and other 3D generation models~\citep{xiang2025structured, li2025triposg}, which makes our method relatively weaker in terms of single-object generation quality, shown in Tab.~\ref{tab:all_comparison}.
\end{reviewmoved}


\end{document}